\documentclass{article}

\usepackage{arxiv}

\usepackage[utf8]{inputenc} 
\usepackage[T1]{fontenc}    
\usepackage{lmodern}        
\usepackage{hyperref}       
\usepackage{url}            
\usepackage{booktabs}       
\usepackage{amsfonts}       
\usepackage{nicefrac}       
\usepackage{microtype}      
\usepackage{graphicx}

\title{Computationally Efficient Learning of Statistical Manifolds}

\author{
    Fan Cheng
   \\
    Monash University \\
  Australia \\
  \texttt{\href{mailto:Fan.Cheng@monash.edu}{\nolinkurl{Fan.Cheng@monash.edu}}} \\
   \And
    Anastasios Panagiotelis
   \\
    University of Sydney \\
  Australia \\
  \texttt{\href{mailto:Anastasios.Panagiotelis@sydney.edu.au}{\nolinkurl{Anastasios.Panagiotelis@sydney.edu.au}}} \\
   \And
    Rob J Hyndman
   \\
    Monash University \\
  Australia \\
  \texttt{\href{mailto:Rob.Hyndman@monash.edu}{\nolinkurl{Rob.Hyndman@monash.edu}}} \\
  }

\usepackage{natbib,amssymb,tabu,placeins,algorithmic}
\usepackage{mathtools,bm}
\usepackage[ruled,vlined,linesnumbered]{algorithm2e}
\usepackage[cal = txupr]{mathalpha}
\mathtoolsset{showonlyrefs}

\begin{document}
\maketitle

\begin{abstract}
Analyzing high-dimensional data with manifold learning algorithms often
requires searching for the nearest neighbors of all observations. This
presents a computational bottleneck in statistical manifold learning
when observations of probability distributions rather than vector-valued
variables are available or when data size is large. We resolve this
problem by proposing a new method for approximation in statistical
manifold learning. The novelty of our approximation is the strongly
consistent distance estimators based on independent and identically
distributed samples from probability distributions. By exploiting the
connection between Hellinger/total variation distance for discrete
distributions and the L2/L1 norm, we demonstrate that the proposed
distance estimators, combined with approximate nearest neighbor
searching, could largely improve the computational efficiency with
little to no loss in the accuracy of manifold embedding. The result is
robust to different manifold learning algorithms and different
approximate nearest neighbor algorithms. The proposed method is applied
to learning statistical manifolds of electricity usage. This application
demonstrates how underlying structures in high dimensional data,
including anomalies, can be visualized and identified, in a way that is
scalable to large datasets.
\end{abstract}


\rhead{Cheng, Panagiotelis, Hyndman: 10 March 2022}

\newpage

\hypertarget{introduction}{%
\section{Introduction}\label{introduction}}

Dimension reduction is an important tool for exploring and analyzing
data for tasks such as clustering, classification, visualization, and
anomaly detection. If high-dimensional data lie on a lower-dimensional
manifold, then manifold learning algorithms (see \citet{Cayton2005-dp},
\citet{Lee2007-wq}, and \citet{Izenman2012-mx} for reviews) can be used
to extract a low dimensional representation of the data. Applications of
manifold learning algorithms include classification and visualization in
prostate cancer gene expression analysis \citep{Witten2011-vq} and
hyperspectral image analysis \citep{Lunga2014-kc}, image reconstruction
during an MRI acquisition \citep{Zhu2018-jw}, and proteomic signaling
network analysis in cancer \citep{Banerjee2019-at}. Of particular
concern in this paper is the case where the manifold of interest is a
\emph{statistical manifold}, by which we mean a manifold whose elements
are probability distributions rather than vector-valued variables (for a
general overview of statistical manifolds, see \citet{Amari2016-hk}).
Examples of statistical manifolds arise in analyzing cell
characteristics in clinical cell cytometry \citep{Carter2009-ti} and
detecting anomalous probability distributions of household electricity
usage \citep{Hyndman2018-ia}. Many popular manifold learning algorithms
require finding nearest neighbors of each observation. When observations
are vector-valued, the Euclidean or Manhattan distance can be used, and
efficient algorithms exist that circumvent the need to compute all
pairwise distances via brute force. However, in the case of statistical
manifolds, such algorithms cannot be exploited, meaning that finding
nearest neighbors poses a computational bottleneck, particularly when
the number of observations is large. To address this problem, we propose
(and prove the consistency of) an estimator for Hellinger (or total
variation) distance that can also be expressed in terms of the Euclidean
(or Manhattan) distance between two vectors. This can be combined with
nearest neighbor algorithms enabling more computationally efficient
learning of statistical manifolds.

Linear methods for dimension reduction date at least as far back as the
development of Principal Components Analysis (see
\citet{Johnstone2009-gf}, \citet{Li2017-ot}, \citet{Fan2019-qv}, and
\citet{Tang2021-qq} for generalization of PCA), while a number of
non-linear methods were developed in the 1960s
\citep{Shepard1962-ac, Shepard1962-ft, Kruskal1964-iv, Kruskal1964-md}.
A further flourishing in the development of non-linear dimension
reduction techniques gained traction after ISOMAP
\citep{Tenenbaum2000-fr} and Local Linear Embedding (LLE)
\citep{Roweis2000-ni} were introduced in the same issue of
\emph{Science}. These methods assume data lie on a manifold and are
hence known as manifold learning techniques. Some algorithms developed
in the early 2000s include, but are not limited to, Laplacian eigenmaps
\citep{Belkin2003-kz}, Hessian LLE \citep{Donoho2003-am}, local tangent
space alignment \citep{Zhang2003-yi}, diffusion maps
\citep{Nadler2006-cm, Coifman2006-no}, and semi-definite embedding
\citep{Weinberger2006-dc}. When it comes to large-scale data,
Nystr\{"o\}m approach and Locally Linear Landmarks are explored but only
in spectral problems
\citep{Drineas2005-zr, Talwalkar2008-fl, Vladymyrov2013-tr}. More
contemporary algorithms including t-SNE \citep{Van_der_Maaten2008-dv}
and UMAP\citep{McInnes2018-xo}, have been shown to be well-suited to the
visualization of many real-world datasets. While these algorithms all
differ, most contain a common step which is to compute a graph of
\(K\)-nearest neighbors. These include ISOMAP, LLE, Laplacian Eigenmaps,
Hessian LLE, t-SNE, and UMAP, which are the six manifold learning
algorithms we consider in this paper.

Computing \(K\) nearest neighbors can be a significant computational
bottleneck especially when the number of observations is large. A naive
way to find the nearest neighbors is to compute pairwise distances
between all observations - an \(O(N^2)\) operation. For some metrics
including Euclidean and Manhattan distance, more efficient solutions
that find the \(K\) nearest neighbors in \(O(Nlog(N))\) time are
available, such as k-d trees \citep{Bentley1975-zo} and Annoy
\citep{Bernhardsson2016-tf}. Popular software implementations of
manifold learning techniques, including the R package \emph{dimRed}
\citep{Kraemer2018-zf} exploit k-d trees algorithms. For applications
with an extremely large number of observations, faster approximate
versions of k-d trees can be used for nearest neighbor search
\citep{Van_Der_Maaten2014-in}. \citet{McQueen2016-xz} provides one such
implementation through the Python package, \emph{megaman}, which
exploits the \emph{FLANN} \citep{Muja2009-de} package for fast neighbor
searching\footnote{approximate nearest neighbors using k-d trees can be
  implemented in R using the C++ ANN library \citep{mount2010-ann} which
  is wrapped in the R package \emph{RANN} \citep{jefferislab2019-l2}}.
However, for statistical manifolds, the distance between observations is
no longer Euclidean, and these approximate nearest neighbor methods are
not applicable.

The main contribution of this paper is to find a way to combine
efficient exact and approximate nearest neighbor algorithms with
learning of statistical manifolds. Previously, \citet{Lee2007-qa}
parameterize discrete probability mass functions as points on a
hypersphere but do not carry out nonlinear dimension reduction,
concluding that \emph{``it is unrealistic in terms of speed to use
algorithms with a complexity of \(O(N^2)\) when \(N\) is large''}.
Recognizing that the approach of \citet{Lee2007-qa} may fail when the
statistical manifold in question lies on a submanifold of the
hypersphere, \citet{Carter2009-ti} propose manifold learning using the
Fisher information metric. However, their method for computing this
metric can be computationally expensive. \citet{Tong2021-ps} develop a
new measuring distance called Diffusion Earth Mover's Distance for large
size of high-dimensional data, but it requires vector-based data and
pre-computed nearest neighbors. Finally, \citet{Hyndman2018-ia} compute
the Jensen Shannon distance between two density estimates and then apply
Laplacian Eigenmaps using the very same dataset we consider in
\autoref{smartmeter}. In all four of these papers, all \(N^2\) pairwise
distances are computed, making them infeasible for applications with
over a few thousand observations.

Estimating the distance (or more generally divergence) between two
probability distributions using samples taken from each distribution is
a well-studied problem in machine learning and statistics. Our focus
will be on non-parametric estimators which make minimal assumptions
about the density functions and can be categorized into plug-in
estimators and direct estimators. In plug-in methods, the estimate of a
distribution function is found using methods such as \(K\)-nearest
neighbors and kernel density estimation. For example,
\citet{Kandasamy2015-iw} consider \(L_2\), Rényi, and Tsallis-\(\alpha\)
divergences via von Mises expansion with plug-in kernel density
estimates. \citet{Poczos2011-wa} focus on the Rényi and Tsallis
divergence estimation. \citet{Poczos2011-wa} and \citet{Poczos2012-ng}
provide Rényi, \(L_2\) divergence, and \(\alpha\)-divergence estimators
using certain \(K\)-nearest neighbors based statistics. Although the
estimator is easy to calculate, the \(K\)-nearest neighbors in the
samples of distributions are pre-required. None of these papers consider
learning of statistical manifolds, and indeed these methods are
impractical for our objective which is to avoid a computationally
expensive \(K\)-nearest neighbor search. For direct methods, a
relationship between the measure function and a functional in the
Euclidean space is found. \citet{Hero2001-pc} and \citet{Hero2002-ik}
investigate the estimation of Rényi divergence but assume that one of
the two density functions is known. \citet{Nguyen2009-fi} and
\citet{Nguyen2010-gc} provide a plug-in estimator for the f-divergence
by estimating the likelihood ratio of two density functions. However,
this involves solving a convex minimization problem, which is very
demanding for large \(N\).

To overcome these issues, we propose an approach similar to that
proposed by \citet{Wang2005-zs}. \citet{Wang2005-zs} estimate KL
divergence by approximating the Radon-Nikodym derivative via a
data-dependent partition of a reference sample space. Since we are
concerned with manifold learning, we require a symmetric distance
measure, and therefore propose an estimate of Hellinger distance or
total variation distance. The reference measure used to partition the
data is the marginal distribution over all \(N\) observational units. By
exploiting a connection between the Hellinger and total variation
distance and the \(L_2\) and \(L_1\) norm respectively, we are able to
find nearest neighbors using k-d trees which has a computational
complexity of \(O(Nlog(N))\) rather than \(O(N^2)\). In principle, even
further speed up may be achieved with approximate nearest neighbor
algorithms. To the best of our knowledge, this approach has not been
previously proposed in the literature and represents the main novel
contribution of our paper. While we acknowledge the shortcomings of
Hellinger and total variation metrics compared to the Fisher information
metric as used by \citet{Carter2009-ti}, we note that Hellinger and
total variation metrics approximate Fisher information metric well
locally; i.e.~when the distance between two probability distributions is
small. This is precisely the situation likely to be encountered when
\(N\) is large and the statistical manifold is more densely sampled.

We demonstrate the potential of our method with an application to smart
meter data where the main objective of the analysis is visualization and
anomaly detection. We show that manifold learning algorithms can be
implemented quickly using k-d trees. The exact version of k-d trees can
even be faster than ANN when the number of grid points used in the
discrete approximation is high. For a fixed manifold learning algorithm,
the effect of using ANN on the accuracy of the embedding and the
identified anomalies is minor. Low dimensional visualizations identify
structure in the data, whether that be in the time of week during which
electricity is used, or in the way that anomalous households from far
away regions of the embedding are very different from one another.

The rest of the paper is organized as follows. In
\autoref{distestimate}, we present our proposed distance estimator and
prove the strong consistency. \autoref{mlann} serves as a primer
defining the algorithms used throughout the paper in detail and with
consistent notation. It is composed of three subsections; the first
deals with different manifold learning algorithms, the second with
approximate nearest neighbor methods, and the last with the quality
measure of embedding used in this paper. Readers familiar with one or
more of these topics may comfortably skip the corresponding subsections.
\autoref{smartmeter} contains the application to visualize and identify
anomalies in household electricity usage using Irish smart meter data.
In this section, we provide further justification for the use of our
distance estimator in statistical manifold learning, including in the
case where the distance between observations is either the Hellinger of
total variation metric between two distributions of electricity usage.
We provide some discussion and conclusions in \autoref{conclusion}.

\hypertarget{distestimate}{%
\section{Distance estimation for statistical manifold
learning}\label{distestimate}}

\hypertarget{estimator}{%
\subsection{Estimator of Hellinger and Total Variation
distance}\label{estimator}}

We now introduce our estimator for the distance between two probability
distributions that can be used when a sample drawn from each
distribution is available. Our emphasis will be on Hellinger distance
which is related to the L2 norm, however, our results extend to the
Total Variation distance which is related to the L1 norm. The Hellinger
distance between two probability measures \(P\) and \(Q\) with sample
space \(\Omega\) is defined as \begin{equation}\label{eq:hellingerdef1}
H(P , Q) = \sqrt{ \frac{1}{2} \int_\Omega \bigg( \sqrt{\frac{dP}{dR}} - \sqrt{\frac{dQ}{dR}} \bigg)^2 dR} ,
\end{equation} where \(P\) and \(Q\) are absolutely continuous with
respect to another measure \(R\), and \(\frac{dP}{dR}\) and
\(\frac{dQ}{dR}\) are the Radon--Nikodym derivatives of \(P\) and \(Q\)
respectively. In the context of manifold learning, \(P\) and \(Q\) will
be the measures associated with a pair of observational units. The
integral above can be approximated by a discrete sum as follows:
\begin{align}
H(P,Q) &\approx \sqrt{\frac{1}{2} \sum_{i=1}^{T} \bigg( \sqrt{\frac{P(I_i)}{R_r(I_i)}} - \sqrt{\frac{Q(I_i)}{R_r(I_i)}} \bigg)^2 R_r(I_i)} \\
&= \sqrt{\frac{1}{2} \sum_{i=1}^{T} \bigg( \sqrt{P(I_i)} - \sqrt{Q(I_i)} \bigg)^2, (\#eq:hellingerest2)}
\end{align} where \(I_1, I_2,\dots, I_T\) are non-overlapping partitions
of the domain of the data (typically \(\mathbb{R}\)). Our estimator
replaces \(\sqrt{P(I_i)}\) and \(\sqrt{Q(I_i)}\) with their empirical
counterparts in a manner that we now describe in detail.

Let \(a_1,\dots,a_m\sim P\) be data corresponding to the first
observational unit and \(b_1,\dots,b_n\sim Q\) be data corresponding to
the second observational unit. For the reference measure \(R\), we will
consider the distribution of the data over all observational units,
\(c_1,\dots,c_r\sim R\). We first find partitions of (empirically) equal
probability using all of the data (i.e.~with respect to R):
\begin{equation}\label{eq:segments}
\{I_i^r \}_{i=1,2,\dots,T_r} = \{ (-\infty, c_{(l_r)}], \ (c_{(l_r)}, c_{(2l_r)} ], \dots, \ (c_{(l_r(T_r-1)}, + \infty ) \},
\end{equation} where \(\{c_{(1)}, c_{(2)}, \dots, c_{(r)}\}\) denote the
order statistics of \(R\),
(i.e.~\(c_{(1)} < c_{(2)} < \dots < c_{(r)}\)),
\(l_r \in \mathbb{N} \leq r\) is the number of sample points from \(R\)
in each segment, and \(T_r\) is the number of segments. Since
\(l_r \in \mathbb{N}\), the last interval contains
\(r-l_r(T_r-1) \leq l_r\) points from \(R\). The quantities
\(\sqrt{P(I_i)}\) and \(\sqrt{Q(I_i)}\) in Equation
@ref(eq:hellingerest2) can be replaced with their empirical counterparts
\(\sqrt{P_m(I^r_i)}:=s_i/m\) and \(\sqrt{Q_n(I^r_i)}:=t_i/n\). Here
\(s_i:=\#\{a:a\in I^r_i\}\) and \(t_i:=\#\{b:b\in I^r_i\}\) are the
number of observations falling into partition \(I_i^r\) from \(P\) and
\(Q\) respectively. The Hellinger distance between \(P\) and \(Q\) can
thus be estimated as \begin{equation}\label{eq:hellingerest1}
\hat{H}(P,Q) =  \sqrt{\frac{1}{2}\sum_{i=1}^{T^r} \left(\sqrt{\frac{s_i}{m}} - \sqrt{\frac{t_i}{n}}\right)^2} \ .
\end{equation}

The right hand side of Equation @ref(eq:hellingerest1) is equivalent to
the Euclidean distance between two vectors
\(\bm{p}:=(s_1/2m,\dots,s_{T^r}/2m)'\) and
\(\bm{q}:=(t_1/2n,\dots,t_{T^r}/2n)'\). In this way, the estimator in
Equation @ref(eq:hellingerest1) allows efficient nearest neighbor
algorithms to be exploited for learning statistical manifolds where only
a sample of data from each distribution on the manifold is available.

The total variation distance between \(P\) and \(Q\) is defined as
\begin{align}\label{eq:tvddef}
TV(P , Q) &= \sup_{E \in \mathcal{F}} |P(E) - Q(E)|
= \frac{1}{2} \int_{\Omega} \left|\frac{dP}{dR} - \frac{dQ}{dR}\right|dR \ ,
\end{align} where \(\Omega\) and \(\mathcal{F}\) are the sample space
and \(\sigma\)-algebra respectively. The estimate for the total
variation distance is derived by similar reasoning as
\begin{equation}\label{eq:tvdest1}
\hat{TV}(P , Q) = \frac{1}{2} \sum_{i=1}^{T_r} \bigg| \frac{s_i}{m} - \frac{t_i}{n} \bigg| ,
\end{equation} which is equal (up to proportionality) to the Manhattan
distance between \(\bm{p}\) and \(\bm{q}\).

\hypertarget{proof-of-convergence}{%
\subsection{Proof of convergence}\label{proof-of-convergence}}

We now prove the consistency as \(n,m,r\rightarrow\infty\) of the
estimator of squared Hellinger distance in Equation
@ref(eq:hellingerest2) and the total variation distance in Equation
@ref(eq:tvddef). The proofs for both estimators are similar, so we will
focus on the proof for the Hellinger distance below, pointing out how
certain steps generalize to the Total Variation where necessary. Our
proof relies on two Lemmas from \citet{Wang2005-zs} which we restate
here in our own notation.

(ref:lemma1caption) Lemma 1 of \citep{Wang2005-zs}

\leavevmode\vadjust pre{\hypertarget{lemma1}{}}%
Let \(R\) be an absolutely continuous probability measure and
\(\{I_i\}_{i=1,\dots,T}\) be a fixed finite partition with
\(I_i = (\gamma_{i-1}, \gamma_i]\) for \(i=1,2,\dots,T\), where
\(-\infty=\gamma_0 < \gamma_1 < \cdots < \gamma_{T-1} < \gamma_T=+\infty\),
such that \(R(I_i)=1/T \text{ for } i=1,2,\dots,T\). Let \(R_r\) be the
empirical probability measure based on i.i.d. samples
\(\{c_1, c_2,\dots, c_r\}\) generated from \(R\) with \(r=l_rT\),
\(l_r\in\mathcal{N}\). Define \(\{I_i^r\}_{i=1,\dots,T}\) as a data
partition where \begin{equation}\label{eq:empsegments}
\{I_i^r \}_{i=1,2,\dots,T} = \{ (-\infty, c_{(1)}^r], \ (c_{(1)}^r, c_{(2)}^r], \dots, \ (c_{(T-2)}^r,c_{(T-1)}^r], (c_{(T-1)}^r, +\infty ) \},
\end{equation} such that \(R_r(I_i^r)=\frac{1}{T}\), for
\(i=1,2,\dots,T\), and where parentheses in subscripts denote order
statistics. Then the sequence of partitions
\(\{\{I_i^r \}_{i=1,2,\dots,T}\}_{r=1,2,\dots}\) converges to the
partition \(\{I_i \}_{i=1,2,\dots,T}\) with respect to \(R\) as
\(r\rightarrow \infty\).

(ref:lemma2caption) Lemma 2 of \citep{Wang2005-zs}

\leavevmode\vadjust pre{\hypertarget{lemma2}{}}%
Let \(R\) be a probability measure with sample space \(\Omega\) and let
\(\{I_1, I_2, \dots, I_T\}\) and
\(\{I_1^r, I_2^r, \dots, I_T^r\}_{r=1,2,\dots}\) be finite measurable
partitions of \(\Omega\). Let \(A\) be an arbitrary probability measure
on \(\Omega\) which is continuous with respect to \(R\). Suppose \(A_m\)
is the empirical probability measure based on i.i.d. samples
\(\{a_1, a_2, \dots, a_m\}\) generated from \(A\). If
\(\{I_1^r, I_2^r, \dots, I_T^r\}_{r}\) converges to
\(\{I_1, I_2, \dots, I_T\}\) with respect to \(R\) as
\(r \rightarrow \infty\), then \begin{equation}
  \lim_{r \rightarrow \infty}\lim_{m \rightarrow \infty} A_m(I_i^r) = A(I_i) \text{ a.s. for } i=1,2,\dots,T.
\end{equation}

We now state our main theorem as follows.

\leavevmode\vadjust pre{\hypertarget{helestimator}{}}%
Let \(P\), \(Q\) be two probability measures each absolutely continuous
with respect to a probability measure \(R\) and let \(H(P,Q)\) and
\(TV(P,Q)\) respectively both be finite, with estimators given by
@ref(eq:hellingerest1) and @ref(eq:tvdest1) respectively. Suppose
\(T_r\) segments with equivalent \(l_r\)-spacing are obtained as in
@ref(eq:segments). If \(T_r, l_r \rightarrow \infty\), then
\begin{equation}
 \hat{H}(P , Q) \rightarrow H(P, Q)
 \qquad \text{and} \qquad
 \hat{TV}(P , Q) \rightarrow TV(P , Q)
\end{equation} as \(m, n,r,T_r \rightarrow \infty\).

The estimation error for the squared Hellinger distance is bounded by
the sum of two terms. \begin{equation}\label{eq:errors}
\begin{aligned}
|\hat{H}^2(P \| Q) - H^2(P \| Q)| &= \frac{1}{2} \bigg| \sum_{i=1}^{T_r} \bigg( \sqrt{\frac{P_m(I_i^r)}{R_r(I_i^r)}} - \sqrt{\frac{Q_n(I_i^r)}{R_r(I_i^r)}} \bigg)^2 R_r(I_i^r) - \int_\Omega \bigg(\sqrt{\frac{dP}{dR}} - \sqrt{\frac{dQ}{dR}} \bigg)^2 dR \bigg| \\
&\leq  \underbrace{\frac{1}{2} \bigg| \sum_{i=1}^{T_r} \bigg( \sqrt{P_m(I_i^r)} - \sqrt{Q_n(I_i^r)} \bigg)^2 -
\sum_{i=1}^{T_r} \bigg( \sqrt{P(I_i)} - \sqrt{Q(I_i)} \bigg)^2 \bigg|}_{e_1} + \\
& \underbrace{\frac{1}{2} \bigg| \sum_{i=1}^{T_r} \bigg(\sqrt{\frac{P(I_i)}{R(I_i)}} - \sqrt{\frac{Q(I_i)}{R(I_i)}} \bigg)^2 R(I_i) - \int_\Omega \bigg(\sqrt{\frac{dP}{dR}} - \sqrt{\frac{dQ}{dR}} \bigg)^2 dR \bigg|}_{e_2}, \\
\end{aligned}
\end{equation} where \(e_1\) is the estimation error induced by the
empirical probability measures on the intervals \(\{I_i\}\), and \(e_2\)
is the approximation error induced by the numerical integration of true
squared Hellinger distance. Both \(e_1\) and \(e_2\) could be proved
below to be arbitrarily small when the sample sizes
\(m, n, \text{and } r\) and the number of points within each interval
\(l_r\) are sufficiently large.

In the case of total variation distance, the estimation error is
decomposed as \begin{equation}\label{eq:tvderrors}
\begin{aligned}
\bigg|\hat{TV}(P , Q) - TV(P , Q) \bigg| &= \frac{1}{2} \bigg| \sum_{i=1}^{T_r} \left| \frac{P_m(I_i^r)}{R_r(I_i^r)} - \frac{Q_n(I_i^r)}{R_r(I_i^r)} \right| R_r(I_i^r) - \int_{\Omega} \left|\frac{dP}{dR} - \frac{dP}{dR}\right|dR \bigg| \\
&\leq \underbrace{\frac{1}{2} \bigg|\sum_{i=1}^{T_r} | P_m(I_i^r) - Q_n(I_i^r) | -
 \sum_{i=1}^{T_r} |P_m(I_i) - Q_n(I_i)| \bigg|}_{e_1} + \\
& \underbrace{\frac{1}{2} \bigg|  \sum_{i=1}^{T_r} \left|\frac{P_m(I_i)}{R_r(I_i)} - \frac{Q_n(I_i)}{R_r(I_i)}\right|R_r(I_i) - \int_{\Omega} \left|\frac{dP}{dR} - \frac{dQ}{dR}\right|dR \bigg|}_{e_2}.
\end{aligned}
\end{equation}

For the approximation error \(e_2\), since the distance between \(P\)
and \(Q\) is finite, we have \begin{equation}
  \frac{1}{2} \int_\Omega \bigg(\sqrt{\frac{dP}{dR}} - \sqrt{\frac{dQ}{dR}} \bigg)^2 dR < \infty,
  \qquad\text{and} \qquad
  \frac{1}{2} \sum_{\omega \in \Omega} \bigg|P(\omega) - Q(\omega)\bigg| < \infty.
\end{equation} Then for any \(\epsilon > 0\), there exists
\(\delta = \delta(\epsilon) > 0\) such that if
\(\max_i{R(I_i)} < \delta\), we have \(e_2 < \frac{\epsilon}{2}\).
Intuitively, this arises since the integral can be approximated
numerically up to an arbitrarily close level of accuracy by the sum.
Recall that the data partition \(\{I_i\}\) is defined in
@ref(eq:segments) and the number of equidistant intervals is \(T_r\),
then the probability on interval \(\{I_i\}\) is \(R(I_i) = 1/T_r\).
Therefore, for any \(T_r \geq 1 / \delta\), we have
\begin{equation}\label{eq:e2}
e_2 < \frac{\epsilon}{2}.
\end{equation}

For the remainder of the proof, assume a fixed \(T \geq T_r(\epsilon)\),
\(T \geq 1 / \delta\) for which @ref(eq:e2) still holds.

For the error \(e_1\), we need to show the convergence of the empirical
probability measure \(R_r\) to the true measure \(R\) on the data
partition \(\{I_i^r\}\) in \(e_1\). By Lemma @ref(lem:lemma1), as
\(r \rightarrow \infty\), the data partition
\(\{\{I_i^r\}_{i = 1,2, \dots, T} \}_r\) converges to the partition
\(\{I_i\}_{i = 1,2, \dots, T}\) with respect to probability measure
\(R\). By Lemma @ref(lem:lemma2), since \(P \ll R\) and \(Q \ll R\), for
any \(i \in \{1,2,\dots,T\}\), we also have \begin{equation}
  \lim_{m \rightarrow \infty}\lim_{r \rightarrow \infty} P_m(I_i^r) = P(I_i)\, a.s. ,
  \qquad \text{and} \qquad
  \lim_{n \rightarrow \infty}\lim_{r \rightarrow \infty} Q_n(I_i^r) = Q(I_i)\, a.s.\label{eq:asconv}
\end{equation}

By the continuous mapping theorem, this implies
\(\sqrt{P_m(I_i^r)}\overset{a.s.}{\rightarrow}\sqrt{P(I_i)}\) and
\(\sqrt{Q_n(I_i^r)}\overset{a.s.}{\rightarrow}\sqrt{Q(I_i)}\). Using
standard results about linear combinations of almost sure convergent
series and the continuous mapping once again implies that
\begin{equation}
\lim_{m \rightarrow \infty}\lim_{n \rightarrow \infty}\lim_{r \rightarrow \infty}\left(\sqrt{P_m(I_i^r)} - \sqrt{Q_n(I_i^r)} \right)^2\rightarrow\bigg( \sqrt{P(I_i)} - \sqrt{Q(I_i)} \bigg)^2\, a.s.
\end{equation} This ensures that for any \(\epsilon > 0\), there exists
\(M_i = m_i(\epsilon, T)\) and \(N_i = n_i(\epsilon, T)\), such that for
any \(m \geq M_i\), \(n \geq N_i\), and for any
\(i \in \{1,2,\dots,T\}\), \begin{equation}
\label{eq:e1function}
\bigg| f(P_m(I_i^r), Q_n(I_i^r)) - f(P(I_i), Q(I_i)) \bigg| 
= \frac{1}{2} \bigg| \bigg( \sqrt{P_m(I_i^r)} - \sqrt{Q_n(I_i^r)} \bigg)^2 - \bigg( \sqrt{P(I_i)} - \sqrt{Q(I_i)} \bigg)^2 \bigg| 
< \frac{\epsilon}{2T} .
\end{equation}

A similar argument can be used in the case of total variation distance,
where the continuous mapping theorem need only be applied once.

Summing over \(i=1,2,\dots, T\) yields \begin{equation}
\frac{1}{2} \bigg| \sum_{i=1}^{T} \bigg( \sqrt{P_m(I_i^r)} - \sqrt{Q_n(I_i^r)} \bigg)^2 -
\sum_{i=1}^{T} \bigg( \sqrt{P(I_i)} - \sqrt{Q(I_i)} \bigg)^2 \bigg| < \frac{\epsilon}{2} \ ,
\end{equation} for \(m \geq \max_i{M_i}\) and \(n \geq \max_i{N_i}\).
Further, since \(T > T_r\) and \(f(x,y) \geq 0\), we have
\begin{equation}\label{eq:e1}
e_1 \leq \frac{1}{2} \bigg| \sum_{i=1}^{T} \bigg( \sqrt{P_m(I_i^r)} - \sqrt{Q_n(I_i^r)} \bigg)^2 -
\sum_{i=1}^{T} \bigg( \sqrt{P(I_i)} - \sqrt{Q(I_i)} \bigg)^2 \bigg| < \frac{\epsilon}{2} \ .
\end{equation}

Combining @ref(eq:e2) and @ref(eq:e1), we have \(e_1 + e_2 < \epsilon\),
hence completing the proof.

\hypertarget{mlann}{%
\section{Algorithms for manifold learning and nearest neighbor
searching}\label{mlann}}

\hypertarget{notation}{%
\subsection{Notation}\label{notation}}

Manifold learning finds a \(d\)-dimensional representation of data that
lie on a manifold \(\mathcal{M}\) embedded in a \(p\)-dimensional
ambient space with \(d \ll p\). We will denote the original data (or
`input' points) as \(x_i\), \(i=1,\dots,N\), where
\(x_i\in\mathbb{R}^p\), while the low-dimensional representation (or
`output' points) will be denoted as \(y_i\), \(i=1,\dots,N\), where
\(y_i\in\mathbb{R}^d\). Where two subscripts are used (e.g.~\(x_{ih}\)
or \(y_{ih}\)), the second subscript refers to the \(h^{th}\) coordinate
or dimension of the data. Pairwise distances between input points
\(x_i\) and \(x_j\) are denoted \(\delta_{ij}\) while pairwise distances
between output points \(y_i\) and \(y_j\) are denoted \(d_{ij}\), where
\(i,j=1,\dots,N\). Unless otherwise stated, these distances are assumed
to be Euclidean. Also important is the \(K\)-ary neighborhoods (or \(K\)
nearest neighbors) of \(x_i\) denoted \(U_K(i)\) and defined as a set of
points \(j\) such that \(x_j\) is one of the \(K\) or less closest
points to \(x_i\).

For the manifold learning algorithms discussed in \autoref{ml}, it is
often assumed that the manifold lies in a \(p\)-dimensional ambient
space. Although the approximate nearest neighbor algorithms we consider
can be applied to metric spaces that are not necessarily Euclidean, they
generally require that observations can be characterized as vectors in
\(\mathbb{R}^p\). Therefore, it may not immediately be clear how these
algorithms can be applied to a statistical manifold, which brings the
main contribution of the paper. The key advantage of estimating the
empirical distributions as discrete over reference bins is that it
allows each distribution to be characterized as a vector. In this way,
the Hellinger distance estimator between the empirical distributions in
\autoref{estimator} would be expressed in \(L_2\) norm of squared root
vectors, making it possible to conduct computationally efficient
manifold learning when the observations are probability distributions.

\hypertarget{ml}{%
\subsection{Manifold learning algorithms}\label{ml}}

We now briefly describe the manifold learning algorithms used in the
remainder of the paper. One feature shared by all algorithms discussed
below is that they require \(K\) nearest neighbors to be found for all
input points. In all implementations of manifold learning algorithms in
\autoref{smartmeter}, \(K\) is set to 20 for comparison across different
methods. Other parameters of the manifold learning algorithms, such as
the intrinsic dimension of the manifold \citep{Denti2021-jl}, are also
essential to the complexity of the data and the embeddings. For
visualization purposes, the dimension \(d\) is set to 2 in this paper.

\hypertarget{isomap}{%
\subsubsection*{ISOMAP}\label{isomap}}
\addcontentsline{toc}{subsubsection}{ISOMAP}

ISOMAP \citep{Tenenbaum2000-fr}, short for isometric feature mapping,
was one of the first algorithms introduced for manifold learning as a
non-linear extension to classical Multidimensional Scaling (MDS) and is
described in Algorithm~@ref(alg:isomap). Classical MDS uses the spectral
decomposition to find an embedding such that the interpoint distances of
the input points \(\delta_{ij}\) and interpoint distances of the output
points \(d_{ij}\) are similar. ISOMAP replaces \(\delta_{ij}\) with
estimates of the geodesic distance between \(x_i\) and \(x_j\) along the
manifold. The geodesic distances are approximated by constructing a
nearest neighbor graph and then finding the shortest path between two
points along this graph using Dijkstra's method \citep{Dijkstra1959-ml}
or Floyd's method \citep{Floyd1962-hx}.

\hypertarget{lle}{%
\subsubsection*{LLE}\label{lle}}
\addcontentsline{toc}{subsubsection}{LLE}

Local Linear Embedding {[}LLE; \citet{Roweis2000-ni}{]} aims to preserve
the local properties of the data and is suited to embedding non-convex
manifolds. Details of LLE are provided in Algorithm~@ref(alg:lle).
First, each input point \(x_i\) is approximated as a linear combination
of its \(K\)-nearest neighbors, \(x_j:j\in U_K(i)\), with \(w_{ij}\)
denoting the weight on \(x_j\) used to approximate \(x_i\). In the next
step, output points are found such that each \(y_i\) is well
approximated by a linear combination of \(y_j:j\in U_K(i)\) and with the
weights \(w_{ij}\) identical to those computed in the previous step.

Analytical solutions exist both for finding the weights and then, given
the weights, to find the output points. The latter involves an
eigendecomposition of a matrix that is sparse due to the fact that only
\(K\) nearest neighbors are used to compute the weights.

\hypertarget{laplacian-eigenmaps}{%
\subsubsection*{Laplacian Eigenmaps}\label{laplacian-eigenmaps}}
\addcontentsline{toc}{subsubsection}{Laplacian Eigenmaps}

Laplacian Eigenmaps \citep{Belkin2003-kz} is based on finding a mapping
\(f\) from the manifold to \(\mathbb{R}^d\) such that the average
squared gradient \(\int_{x\in\mathcal{M}}||\nabla f(x)||^2\) is
minimized. The rationale is that input points close to one another on
the manifold should be mapped to output points that are close to one
another, hence \(f\) should be as flat as possible. Minimizing the
average squared gradient of \(f\) corresponds to finding eigenfunctions
of the Laplace-Beltrami operator. In practice, eigenvectors of the graph
Laplacian are found where the graph Laplacian serves as a discrete
approximation to the Laplace-Beltrami operator. The graph Laplacian is
computed from the \(K\)-nearest neighbors graph in a way that is
described in Algorithm~@ref(alg:le).

\hypertarget{hessian-lle}{%
\subsubsection*{Hessian LLE}\label{hessian-lle}}
\addcontentsline{toc}{subsubsection}{Hessian LLE}

Hessian LLE (HLLE; \citet{Donoho2003-am}) works with the functional
\(\mathcal{H}(f)=\int_{x\in\mathcal{M}}||H(f(x))||_F\) which is the
average Frobenius norm of the Hessian of a function \(f\) where
\(f:\mathcal{M}\rightarrow\mathbb{R}^d\). The Hessian is defined using
orthogonal coordinates on the tangent space of \(\mathcal{M}\) which are
approximated by taking a singular value decomposition of the
\(K\)-nearest neighbors around each point. If there exists a mapping
from the manifold \(\mathcal{M}\) to \(\mathbb{R}^d\) that is locally
isometric, then the null space of \(\mathcal{H}(f)\) is spanned by the
original coordinates. The functional \(\mathcal{H}(f)\) can be
estimated, and spectral methods are then used to find the null space and
hence recover the isometric coordinates. This is outlined in
Algorithm~@ref(alg:hlle).

\hypertarget{t-sne}{%
\subsubsection*{t-SNE}\label{t-sne}}
\addcontentsline{toc}{subsubsection}{t-SNE}

t-Distributed Stochastic Neighbor Embedding {[}t-SNE;
\citet{Van_der_Maaten2008-dv}{]} is a dimension reduction technique well
suited for the visualization of high-dimensional data in scatterplots.
It works by minimizing the Kullback-Leibler divergence
\(K L(P \| Q)=\sum_{i} \sum_{j} p_{ij} \log \frac{p_{ij}}{q_{ij}}\)
between two distributions, \(p_{ij}\) and \(q_{ij}\), respectively
giving transition probabilities between points in the input and outpoint
space. To ensure transitions to nearby points are more probable, the
transition probabilities \(p_{ij}\) are proportional to a Gaussian
kernel evaluated at the Euclidean distance \(\delta_{ij}\). To better
capture the local data structure and solve the crowding
problem\citep{Van_der_Maaten2008-dv} for nearby points, \(q_{ij}\) are
chosen to be proportional to \(d_{ij}\) evaluated at a normalized
Student-t kernel. The bandwidth of the Gaussian kernel, \(\sigma_i\) is
determined by the \emph{perplexity} which is set to 30 by default. An
analytical form of the gradient is used in the Gradient Descent
optimization of Kullback-Leibler divergence with details described in
Algorithm~@ref(alg:tsne). Variants of the Barnes-Hut algorithm can also
be used to approximate the gradient of the object function
\citep{Van_Der_Maaten2014-in}.

\hypertarget{umap}{%
\subsubsection*{UMAP}\label{umap}}
\addcontentsline{toc}{subsubsection}{UMAP}

UMAP {[}Uniform Manifold Approximation and Projection;
\citet{McInnes2018-xo}{]} preserves global structure is motivated by
Riemannian geometry and algebraic topology with a large computational
advantage over t-SNE. Rather than minimizing interpoint transition
probabilities as in t-SNE, UMAP minimizes the error between two
topological representations from the input space and embedding space.
The error is measured by the cross entropy \(C\) of two local fuzzy
simplicial sets built as weighted nearest neighborhood graph, with edge
weights representing the likelihood that two points are connected.
Geometrically, a simplex is a \(K\)-dimensional object built from
connecting \(K+1\) points. UMAP captures topological structure by
treating the data as a set of simplices and combining them in a specific
way described in Algorithm~@ref(alg:umap). In order to connect with
nearby points, UMAP extends a radius outwards at each point until these
overlap. The desired separation between nearby embedded points is
controlled by the parameter \textit{min\_dist}. Finally, after
initializing the embedding using the weighted graph Laplacian,
stochastic gradient descent is used to find the most similar graph in
lower dimensions. Further details of the algorithm and hyperparameters
can be found in Section 4 of \citet{McInnes2018-xo}.

\hypertarget{ann}{%
\subsection{Approximate nearest neighbor searching}\label{ann}}

\begin{figure}

{\centering \includegraphics[width=0.8\linewidth]{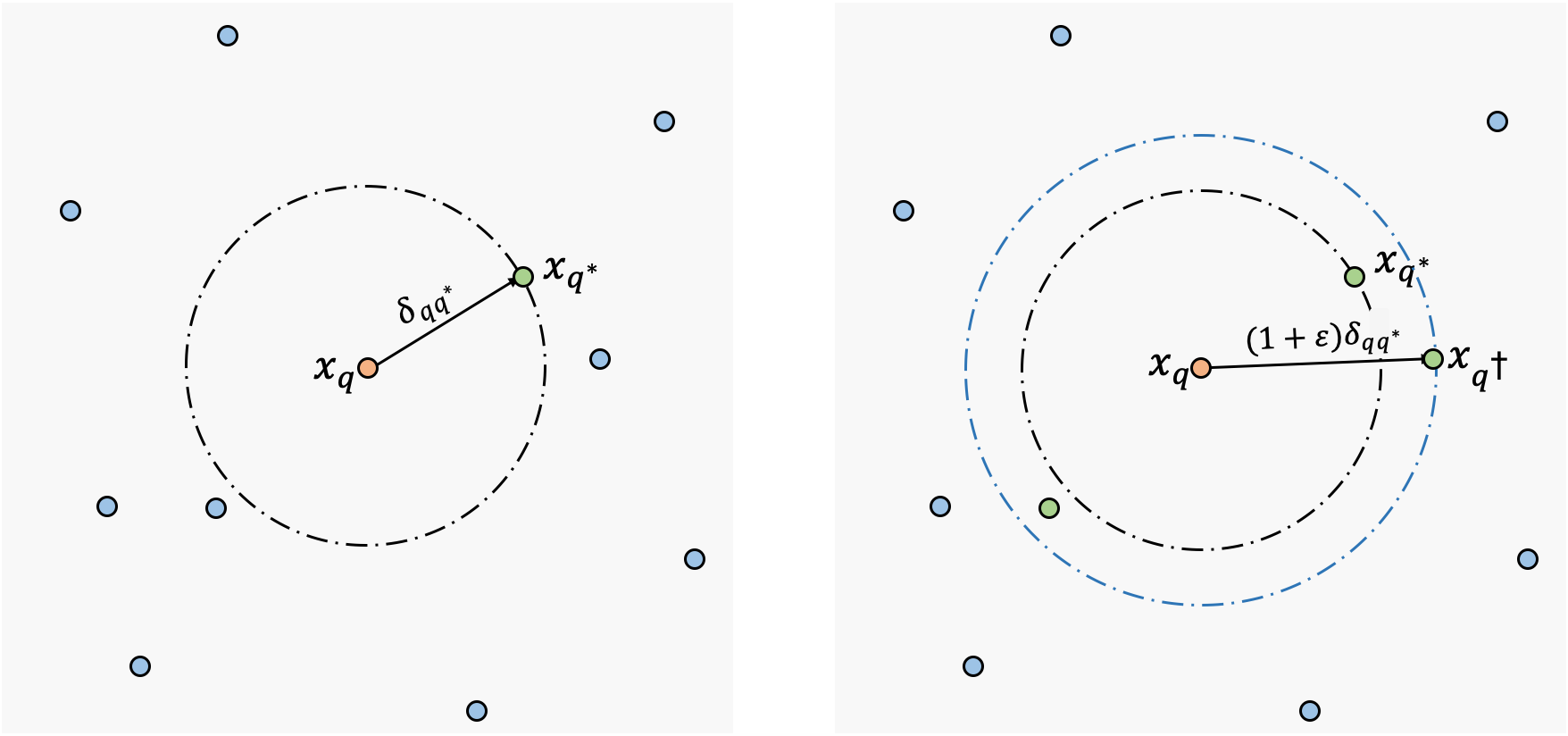} 

}

\caption{Illustration of approximate nearest neighbors (right) compared to exact nearest neighbors (left). The left subplot shows the distance from the true nearest neighbor point $x_{q^*}$ to the query point $x_q$ is $\delta_{q{q^*}}$, while the right subplot shows that the $(1+\varepsilon)$ -approximate nearest neighbors (green points) lie within $(1+\varepsilon) \delta_{q{q^*}}$ radius from $x_q$. }\label{fig:ann}
\end{figure}

As seen in the previous section, many manifold learning algorithms begin
by finding the graph of nearest neighbors. A naive way to find this
graph is to calculate all pairwise distances between each pair of the
input data points. This has a complexity of \(O(N^2)\) for \(N\)
observations, which is not efficient when the data set is large.
Although faster algorithms exist for exact nearest neighbors, for very
large \(N\) we propose the use of approximate nearest neighbor
algorithms. Whereas an exact nearest neighbor algorithm will, for a
given query point \(x_q\), return the point \(x_{q^*}\) such that
\(\delta_{q{q^*}} \leq \delta_{qj}\) for all \(j\neq q^*\), an
approximate nearest neighbor algorithm returns a point \(x_{q^\dagger}\)
such that \(\delta_{q{q^\dagger}} \leq (1+\varepsilon) \delta_{q{q^*}}\)
for some tolerance level \(\varepsilon > 0\) \citep{Arya1998-bv}. This
is illustrated in \autoref{fig:ann}. This definition can be generalized
to \(K\)-nearest neighbors. Since our objective is to find the \(K\)
nearest neighbor graph, our query points are the points in the original
sample. So the nearest neighbor of each \(x_q\) will be \(x_q\) itself.
This is disregarded and we search for \(K+1\) nearest neighbors.

\citet{Aumuller2020-nk} developed a benchmark tool\footnote{available at
  \url{http://ann-benchmarks.com/}.} and evaluated a number of
contemporary approximate nearest neighbor searching methods, among which
Annoy is one of the competitive methods. We will also use k-d trees
since this is one of the most widely used algorithms for (approximate)
nearest neighbor search. In addition, k-d trees also allow for the
special case of exact nearest neighbor search through appropriate
selection of the tuning parameters, thus easily allowing us to isolate
the effect of using approximate nearest neighbors in manifold learning
algorithms. We now briefly describe the approximate nearest neighbor
algorithms used in our evaluation, namely, k-d trees and Annoy.

\hypertarget{k-d-trees}{%
\subsubsection*{k-d trees}\label{k-d-trees}}
\addcontentsline{toc}{subsubsection}{k-d trees}

The k-d tree \citep{Bentley1975-zo}, is a binary tree structure that can
be exploited for nearest neighbor search. Each node is associated with a
partition of \(p\)-dimensional space \(\mathcal{P}_g\) (hereafter the
node and partition will be treated interchangeably). A node has two
children --- a left child node and a right child node --- formed by
splitting \(\mathcal{P}_g\) by an axis-orthogonal hyper-plane. By a
variant of k-d trees proposed by \citet{Friedman1977-dh}, the splitting
dimension \(l_g^*\) for the splitting hyperplane is chosen to maximize
spread. The splitting value \(c_g^*\) is chosen as the median along the
splitting dimension of all points in \(\mathcal{P}_g\). The process of
building a k-d tree is shown in Algorithm~@ref(alg:buildkdtree). Note
that while Algorithm~@ref(alg:buildkdtree) will construct a tree with
terminal nodes that include no more than one data point, variants of k-d
trees allow for multiple points in the terminal nodes.

Once a k-d tree is constructed, nearest neighbors can be searched in
time proportional to \(O(\log(N))\). Nearest neighbor search begins at
the root node and determines whether the query point belongs to the
partition of the left child node or right child node. The search moves
to the child node to which it belongs and this process is iterated until
a terminal node \(\mathcal{P}_T\) is reached. The distance from the
query point to \(x_i\in\mathcal{P}_T\) is set to the current shortest
distance \(\delta^*\) and \(x_i\) to the current nearest neighbor. The
search algorithm then unwinds back through the tree. At each parent
node, there will be one sibling node that has been searched and one
sibling node yet to be searched. Denote the latter as
\(\mathcal{P}_{g^*}\). The tightest bounding box containing all
\(x_i:x_i\in\mathcal{P}_{g^*}\) is computed as well as a hyper-sphere of
radius \(\delta^*\) around the query point. If the hyper-sphere and
bounding box do not intersect, then the nearest neighbor cannot lie in
\(\mathcal{P}_{g^*}\) and the branch starting from this node can be
disregarded. Otherwise, this branch must also be searched. Pseudocode
for a recursive procedure used to find the nearest neighbor \(x_{q^*}\)
of a query point \(x_q\) is given in Algorithm~@ref(alg:searchkdtree).
This search is initialized by evaluating this procedure at the root node
and initializing the current closest distance at \(\delta^*=\infty\).

The algorithm described above will find exact nearest neighbors with
\(O(\log(N))\) operations on average. However, if approximate rather
than exact nearest neighbors are desired, further speed up can be
achieved by setting the radius of the hypersphere at Line 11 and Line 18
of Algorithm~@ref(alg:searchkdtree) to \(\delta^*/(1+\varepsilon)\).
This increases the number of branches of the tree eliminated from the
search. In subsequent sections, we tune \(\varepsilon\) to trade off
between speed and accuracy. In all cases below where exact nearest
neighbors are computed, we employ k-d trees with \(\varepsilon=0\). We
also note that while Algorithm~@ref(alg:searchkdtree) finds the nearest
neighbor, it can be generalized to finding \(K\) nearest neighbors by
replacing the current smallest distance and current closest node with
the current \(K\)-th smallest distance and current \(K\)-th closest node
respectively.

\hypertarget{annoy}{%
\subsubsection*{Annoy}\label{annoy}}
\addcontentsline{toc}{subsubsection}{Annoy}

\begin{figure}

{\centering \includegraphics[width=.49\linewidth]{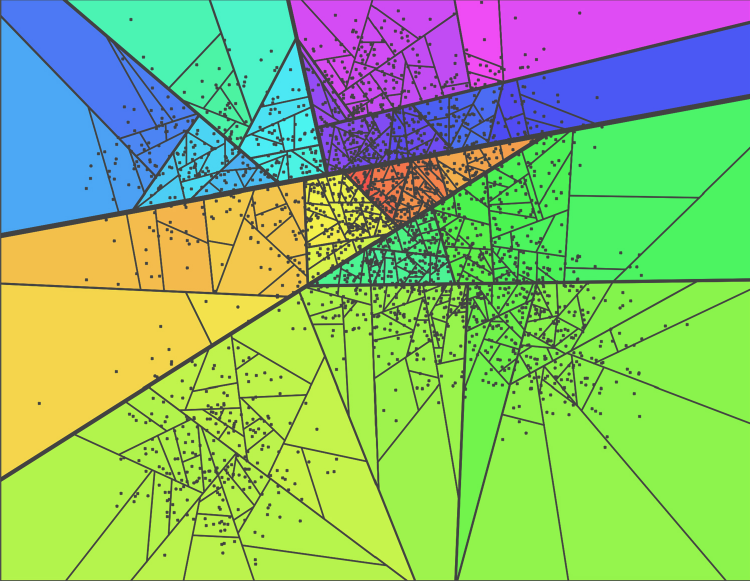} \includegraphics[width=.49\linewidth]{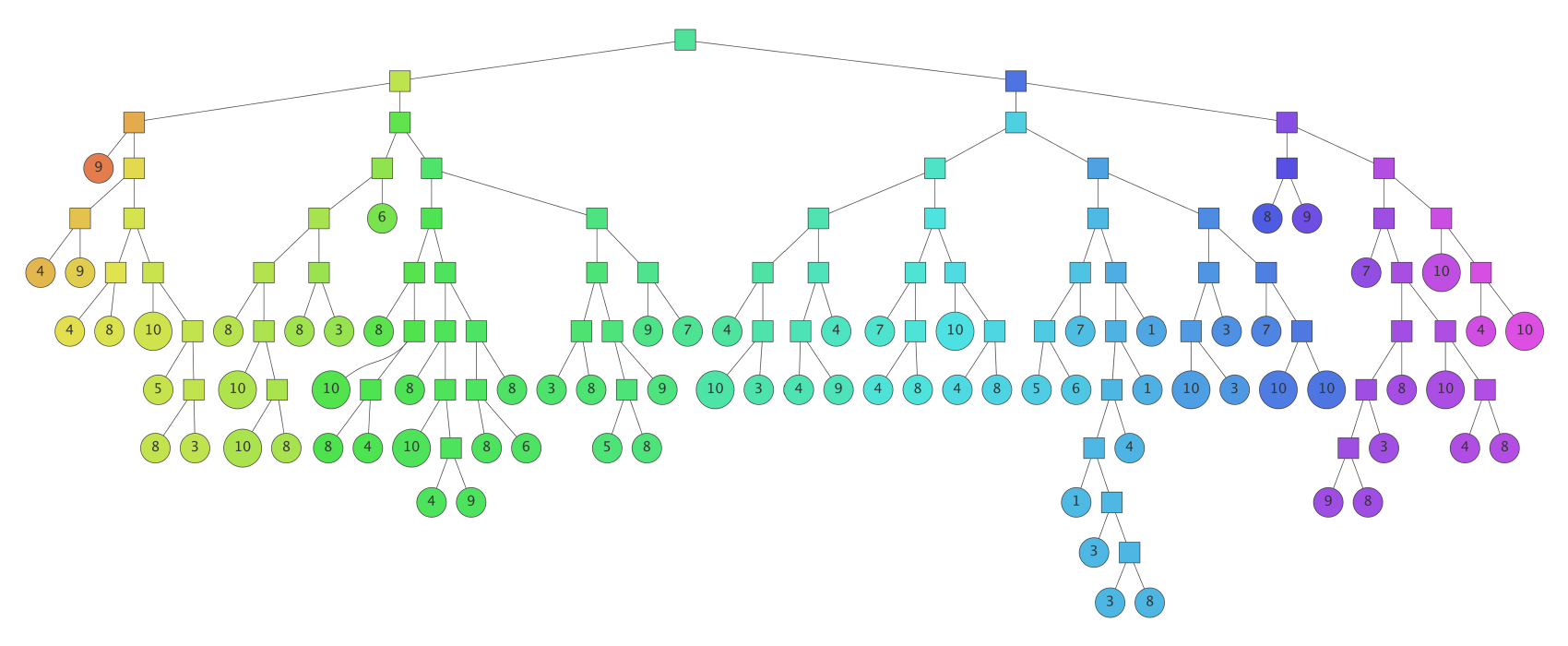} 

}

\caption{(ref:annoycaption)}\label{fig:annoy}
\end{figure}

Annoy {[}Approximate Nearest Neighbors Oh Yeah;
\citet{Bernhardsson2016-tf}{]} is a C++ library with Python bindings
that implements an alternative tree-base algorithm for nearest neighbor
search. While sharing some similarities, Annoy differs from k-d trees in
two main ways. First, rather than using a single tree, the Annoy
algorithm constructs a forest of randomly constructed trees that can be
searched in parallel. Second, while Annoy builds trees by constructing
splitting hyperplanes, unlike k-d trees, these hyperplanes are not
axis-orthogonal. Instead, for each split, Annoy chooses two points at
random and sets the splitting hyperplane to be equidistant from these
two points. Each node (including terminal nodes) is associated with a
partition containing multiple (at most \(\kappa\)) points. One such tree
is shown in \autoref{fig:annoy}.

(ref:annoycaption) The splitting process of Annoy (on the left) and the
corresponding binary tree (on the right) reproduced from
\citet{Bernhardsson2015-slides}.

Annoy searches for nearest neighbors on all trees in the forest in
parallel. The search for nearest neighbor candidates on a single tree
does share some similarities with nearest neighbor search on k-d trees.
In particular, the search begins from the root node and iteratively
moves to the child node associated with a partition containing the query
point. Once a terminal node is reached, points within this node are
considered as nearest neighbor candidates. However, Annoy does not
unwind each tree in the same fashion as k-d trees. Instead, Annoy
maintains a priority queue (shared across all trees of the forest) of
the splitting hyperplanes closest to the query point. If the distance
from the query point to a splitting hyperplane is less than some
threshold \(\eta\), then the opposite side of the hyperplane is also
searched. Increasing the threshold throughout allows more points to be
considered as nearest neighbor candidates.

Once the union of candidate points across all trees contains
\textit{search\_k} points, searching the forest ends. Distances are
computed to each point in the candidate sets and approximate nearest
neighbors found. The accuracy-performance trade-off is controlled by two
parameters, the number of trees, \textit{n\_trees}, and the size of the
candidate set, \textit{search\_k}. In both cases, larger values will
improve the accuracy of nearest neighbor search at the cost of slower
performance. Choosing a larger value of \textit{n\_trees} also increases
the demand for storage. A full description of the Annoy is provided in
Algorithm~@ref(alg:annoypre) and @ref(alg:annoyquery).

\hypertarget{quality-measures-for-manifold-learning-embedding}{%
\subsection{Quality measures for manifold learning
embedding}\label{quality-measures-for-manifold-learning-embedding}}

To examine the effect of using approximate nearest neighbor algorithms
in manifold learning, measures of the quality of a low-dimensional
representation must be defined. \citet{Arias-Castro2020-th} derive
performance bounds for some manifold learning methods based on the study
Perturbation bounds. In the ANN benchmark tool built by
\citet{Aumuller2020-nk}, recall rate is used to measure the accuracy of
the approximate nearest neighbor searching methods. Recall rate is
defined as the proportion of observations correctly classified as
nearest neighbors compared to the true ones. However, even with a large
proportion of true nearest neighbors, the local structure in the
manifold learning embeddings might still be inaccurate. Therefore, the
quality measures of the embeddings are essential. \citet{Venna2006-nd}
defined two quality measures, trustworthiness and continuity, that
respectively distinguish two types of errors. For the first type of
error, \(y_j\) is among the \(K\)-nearest neighbors of \(y_i\)
(i.e.~observations close in output space) but \(x_j\) is not among the
\(K\)-nearest neighbors of \(x_i\) (i.e.~observations not close in the
input space). Using all such points for each \(i\), the trustworthiness
of the embedding can be calculated as
\begin{equation}\label{eq:trustworthiness}
  M_{T}(K)=1-\frac{2}{G_{K}} \sum_{i=1}^{N} \sum\limits_{\begin{subarray}{c}j \in V_{K}(i)\\j \notin U_{K}(i)\end{subarray}}(\rho_{ij}-K),
\end{equation} where the normalizing factor is \begin{equation}
  G_{K}=\left\{
  \begin{array}{ll}
    N K(2 N-3 K-1) & \text { if } K < N / 2, \\
    N(N-K)(N-K-1) & \text { if } K \geq N / 2.
  \end{array}
\right.
\end{equation} This is bounded between zero and one, with values closer
to one indicating a higher-quality representation. In contrast to recall
rate, Trustworthiness uses information on the rankings of interpoint
distances. In particular, points that are incorrectly included as
nearest neighbors in the output space are penalized more when they are
further away in the input space (\(\rho_{ij}\) is high).

\hypertarget{smartmeter}{%
\section{Application to smart meter data}\label{smartmeter}}

\hypertarget{irish-smart-meter-dataset}{%
\subsection{Irish smart meter dataset}\label{irish-smart-meter-dataset}}

Next, we consider smart-meter data for residential and non-profiled
meter consumers, collected in the \emph{CER Smart Metering Project -
Electricity Customer Behaviour Trial, 2009-2010} in Ireland
\citep{cer2012-data}. Electricity smart meters record consumption, on a
near real-time basis, at the level of individual commercial and
residential properties. The CER dataset\footnote{accessed via the Irish
  Social Science Data Archive - www.ucd.ie/issda.} does not include
energy for heating systems since it is either metered separately, or
households use a different source of energy, such as oil or gas. In this
study, the installed cooling systems are also not reported.

\begin{figure}

{\centering \includegraphics[width=0.8\linewidth]{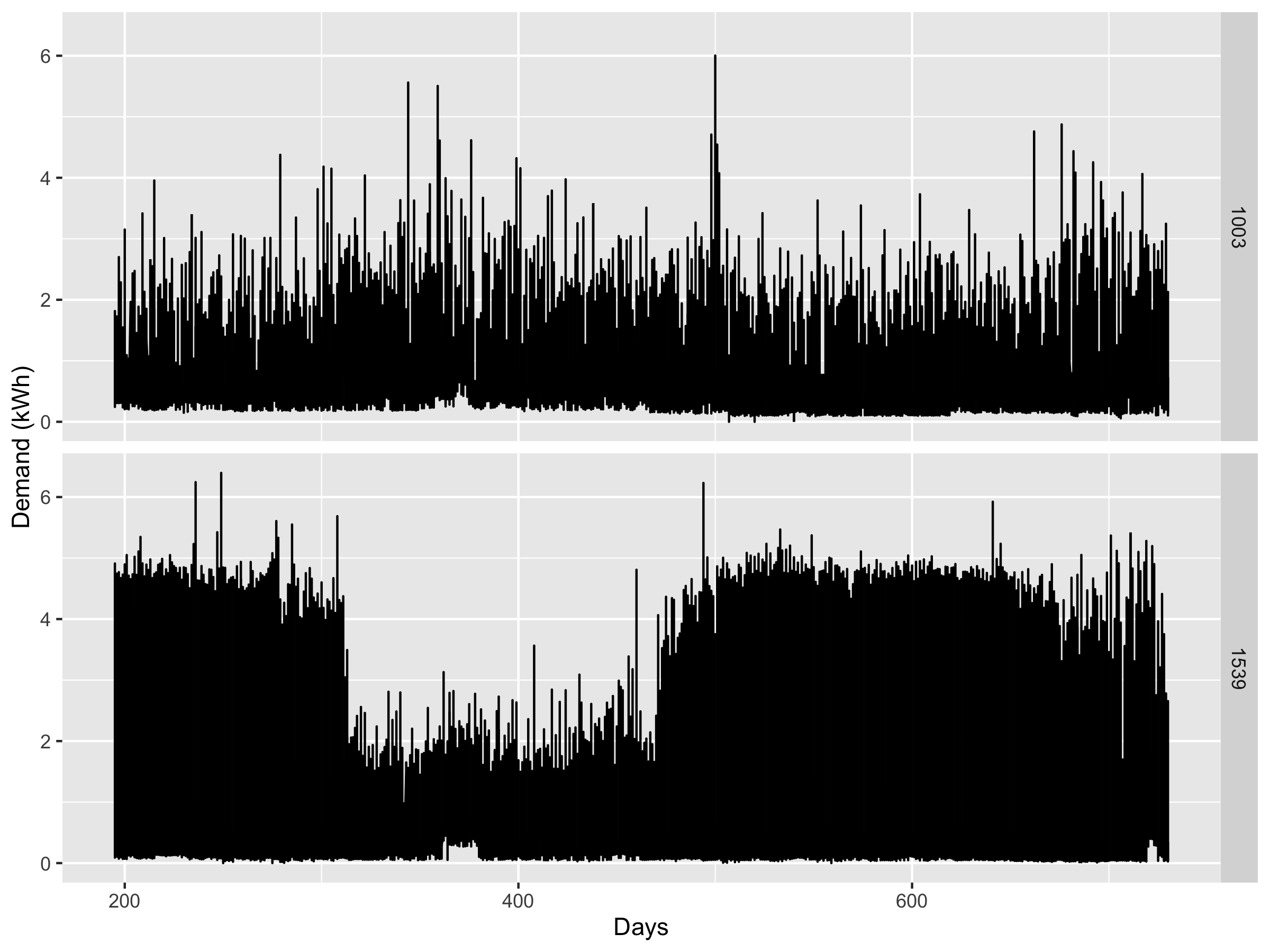} 

}

\caption{Two smart-meter demand examples, ID 1003 and ID 1539, from the Irish smart meter data set.}\label{fig:smartmeter}
\end{figure}

We use measurements of half-hourly electricity consumption gathered from
3,639 residential consumers over 535 consecutive days. Every meter
provides electricity consumption between 14 July 2009 and 31 December
2010. Demand data from two smart meters (ID 1003 and 1539) are shown in
\autoref{fig:smartmeter} as time series plots. It is obvious that these
meters have relatively different patterns. Meter 1539 (bottom of
\autoref{fig:smartmeter}) has a period of around 150 days with lower
(approximately half) the electricity usage of remaining days and is
otherwise relatively stable. In contrast, Meter 1003 (top of
\autoref{fig:smartmeter}) exhibits regular spikes on weekends.

For electricity demand data, one particular problem of interest is to
visualize and identify households or periods of the week with anomalous
usage patterns. For this reason, the object of interest in comparing
households or periods of the week is the distribution of electricity
demand rather than the raw data itself \citep{Hyndman2018-ia}. An
additional advantage of looking at distributions rather than raw data is
that it provides a convenient mechanism for handling missing data which
can be a significant problem in smart meter data applications. In the
rest of this section, we first describe how discrete approximations to
the distributions of interest are computed. We then approximate the
distances between vectors containing probabilities with the consistent
estimator in \autoref{estimator}, which allows us to apply manifold
learning techniques on statistical manifolds.

\hypertarget{dataprocessing}{%
\subsection{Data processing}\label{dataprocessing}}

Let \(d_{i,t}\) denote the electricity demand for observation \(i\) and
for time period \(t\) (subsequently we will see that \(i\) can either
index the household, time of the week, or both, while \(t\) may index
the week or half-hour period). The objective is to estimate the
distances between the distribution of electricity demand for observation
\(i\), \(F_i\), over time. Since the raw data could be thought of as
samples generated from the true electricity usage distributions for
different households and different periods, we could apply the
consistent Hellinger distance estimator derived in \autoref{estimator}
to the smart meter data. The first step is to group all data together
regardless of \(i\) and \(t\), and take the data as samples from the
reference probability distribution \(R\). The sample size
\(r=93,616,914\). Next, we construct \(T_r=100\) intervals with about
\(1\%\) samples from \(R\) within each interval (except for the last
interval). The segments \(\{I_g^r\}_{g=1,2,\dots,100}\) can then be used
to construct discrete distribution approximation to \(F_i\) by counting
the ratio of samples points from \(F_i\) fallen into segment \(I_g^r\).
This is found by computing
\(\pi_{i,g}=(1/m)\sum_t I(d_{i,t} \in I_g^r)\) where \(m\) is the sample
size from \(F_i\). Then the Hellinger distance between \(F_i\) and
\(F_j\) could be estimated using Equation @ref(eq:hellingerest1). In
this way, we make it possible to search nearest neighbors more
efficiently using ANN methods and further apply manifold learning
algorithms.

\hypertarget{electricityresults}{%
\subsection{Manifold learning results for single
household}\label{electricityresults}}

\begin{figure}[!b]

{\centering \includegraphics[width=1\linewidth]{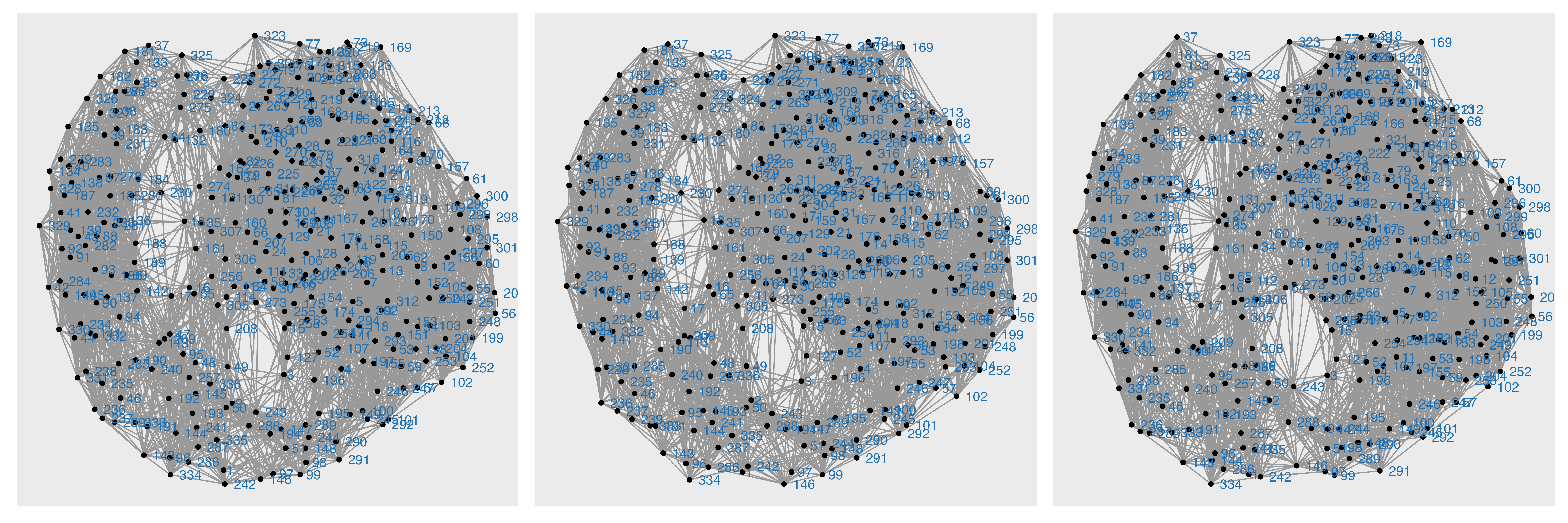} 

}

\caption{Nearest neighborhood graphs for meter ID 1003 with $K=20$. The left subplot is the exact nearest neighborhood graph, while the middle and right subplots are the approximate ones using k-d trees ($\varepsilon = 1$) and Annoy ($\textit{n\_trees = search\_k}=50$) respectively.}\label{fig:neighborplot}
\end{figure}

To start with, we consider the case of a single household where each
observation is a distribution corresponding to a single half-hour of the
week so that each \(i\) corresponds to one of \(48 \times 7 = 336\)
day-time pairs and \(t\) corresponds to a week. In this case, \(N=336\)
in the manifold learning. We choose meter ID 1003 for this purpose since
it is a household with a fairly typical pattern of electricity usage.
\autoref{fig:neighborplot} shows a nearest neighbors graphs with
\(K=20\) for meter ID 1003. The left two panels are found using k-d
trees with the left panel being exact (\(\varepsilon = 0\)) and the
middle panel being approximate (\(\varepsilon = 1\)). We note that the
ANN output with \(\varepsilon = 1\) does not differ much from other
values such as a \(\varepsilon\) of 0.5 or 2. We also use Annoy with
\(\textit{n\_trees}=\textit{search\_k}=50\) to plot the corresponding
nearest neighbor graph on the right panel of \autoref{fig:neighborplot}.
The three subplots appear quite similar to each other with a recall rate
of 1 for the exact graph, 0.978 for k-d trees, and 0.961 for Annoy,
suggesting that for the smart meter data, ANN techniques can be used
within manifold learning to save computation time and memory without
having too severe an impact on the quality of the low-dimensional
representation.

\begin{figure}

{\centering \includegraphics[width=1\linewidth]{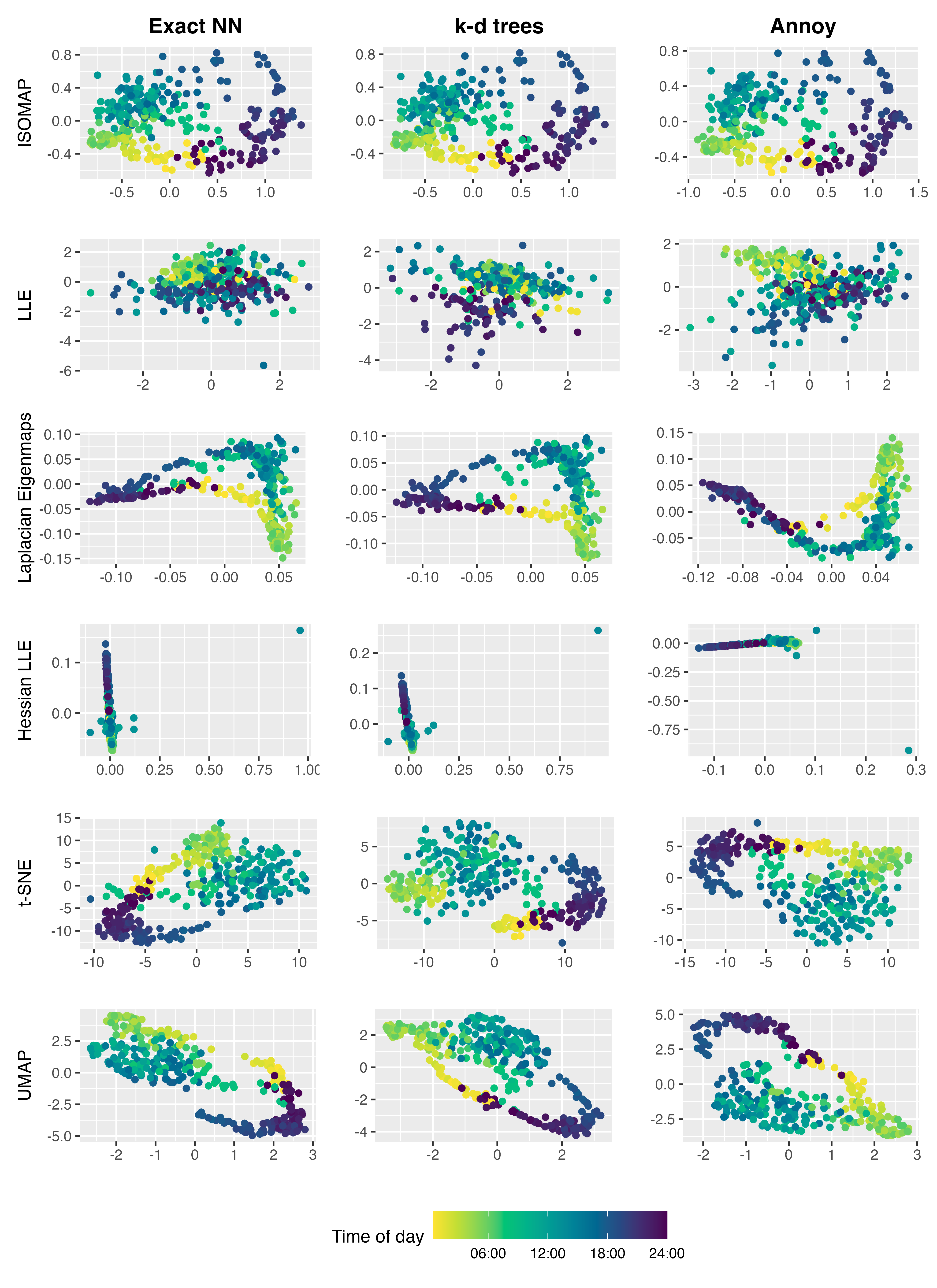} 

}

\caption{Embeddings from six manifold learning methods (ISOMAP, LLE, Laplacian eigenmaps, Hessian LLE, t-SNE and UMAP) for one household from the smart meter data. Each subplot is a scatterplot of the 2-d embedding points, with the color representing half-hourly time of the day. Left panel: exact nearest neighbors. Middle panel: ANN with k-d trees. Right panel: ANN with Annoy.}\label{fig:todplot}
\end{figure}

Since a key objective is the visualization of anomalous times of the
week, \autoref{fig:todplot} shows \(d=2\)-dimensional representations of
the data for meter ID 1003 obtained using, from left to right, ISOMAP,
LLE, Laplacian Eigenmaps, Hessian LLE, t-SNE and UMAP. All cases in the
top panels refer to results when exact nearest neighbors are used, while
the middle and bottom panels refer to results where approximate nearest
neighbors are used. The half-hour of the day that an observation belongs
to is depicted using color. With the exception of LLE and Hessian LLE,
the results demonstrate that similar times of day are grouped closely
together. The cyclical pattern observed in the representation is
indicative of the fact that low values for half-hour of the day (00:00,
00:30, 01:00) are temporally proximate to high values for half-hour of
the day (22:30, 23:00, 23:30) and are therefore similar.
\autoref{fig:todplot} also exhibits three clusters roughly corresponding
to three phases of a typical day, working during the day (08:00 --
17:00), recreation during the evening (17:00 -- 00:00), and sleeping
(00:00 -- 08:00). The times of day were not explicitly used in
constructing results, rather this structure was uncovered by the
manifold learning algorithms themselves. This grouping is particularly
prominent in tSNE and UMAP while to a lesser extent in Laplacian
Eigenmaps. Encouragingly, when comparing row-wise, all patterns are
equally prominent irrespective of whether exact or approximate nearest
neighbors are used.

\autoref{tab:quality1id} shows one of the embedding quality measures,
Trustworthiness, for all the embeddings in \autoref{fig:todplot}. By
comparing the Trustworthiness, it can be shown that using ISOMAP gives
the highest Trustworthiness of 0.943, while k-d trees and Annoy lead to
almost the same trustworthiness as even exact nearest neighbors. This
conclusion can also be drawn for almost all the other quality measures
mentioned in the online supplementary document. The running times are
summarized in \autoref{tab:time1id}, with the run time for k-d trees
generally shorter than Annoy. Note that we use the same k-d trees
implementation for exact nearest neighbor searching, but with
\(\varepsilon=0\). For comparison, we include the running time for
brute-force method, which is slower than both the k-d trees and Annoy
implementations. In general, the improvement from using approximate
nearest neighbors relative to exact nearest neighbors is fairly minor in
this small subset of the data. However, we reiterate that even exact
nearest neighbors here use k-d trees, compared to the brute-force
methods in the literature that require all \(N^2\) pairwise distances to
be computed and are thus infeasible.

\begin{table}

\caption{\label{tab:quality1id}Comparison of Trustworthiness measure using true nearest neighbors, k-d trees and Annoy for six manifold learning methods in meter ID 1003. ANN give almost the same or higher Trustworthiness than exact nearest neighbors. }
\centering
\resizebox{\linewidth}{!}{
\begin{tabu} to \linewidth {>{\raggedright}X>{}r>{}r>{}r>{}r>{}r>{}r}
\toprule
  & Isomap & LLE & Laplacian Eigenmaps & Hessian LLE & t-SNE & UMAP\\
\midrule
Exact NN & \textbf{0.943} & 0.709 & 0.885 & 0.882 & 0.935 & 0.924\\
ANN k-d trees & 0.942 & 0.718 & \textbf{0.891} & \textbf{0.885} & 0.902 & \textbf{0.933}\\
ANN Annoy & 0.938 & \textbf{0.746} & 0.886 & 0.873 & \textbf{0.939} & 0.929\\
\bottomrule
\end{tabu}}
\end{table}

\begin{table}

\caption{\label{tab:time1id}Comparison of computation time using true nearest neighbors, k-d trees, and Annoy for six manifold learning methods in meter ID 1003. Generally, exact NN with brute-force is slower than k-d trees implementations for both exact and approximate NN and Annoy. }
\centering
\resizebox{\linewidth}{!}{
\begin{tabu} to \linewidth {>{\raggedright\arraybackslash}p{5cm}>{}r>{}r>{\raggedleft\arraybackslash}p{2cm}>{}r>{}r>{}r}
\toprule
  & Isomap & LLE & Laplacian Eigenmaps & Hessian LLE & t-SNE & UMAP\\
\midrule
Exact NN with brute-force & \textbf{2.760} & \textbf{5.134} & \textbf{2.944} & \textbf{2.087} & \textbf{1.918} & \textbf{2.367}\\
Exact NN with k-d trees & 2.360 & 2.609 & 2.295 & 1.669 & 1.596 & 2.089\\
ANN k-d trees & 2.351 & 2.986 & 2.302 & 1.663 & 1.601 & 2.048\\
ANN Annoy & 1.942 & 3.093 & 2.378 & 1.695 & 1.647 & 2.125\\
\bottomrule
\end{tabu}}
\end{table}

\begin{figure}

{\centering \includegraphics[width=1\linewidth]{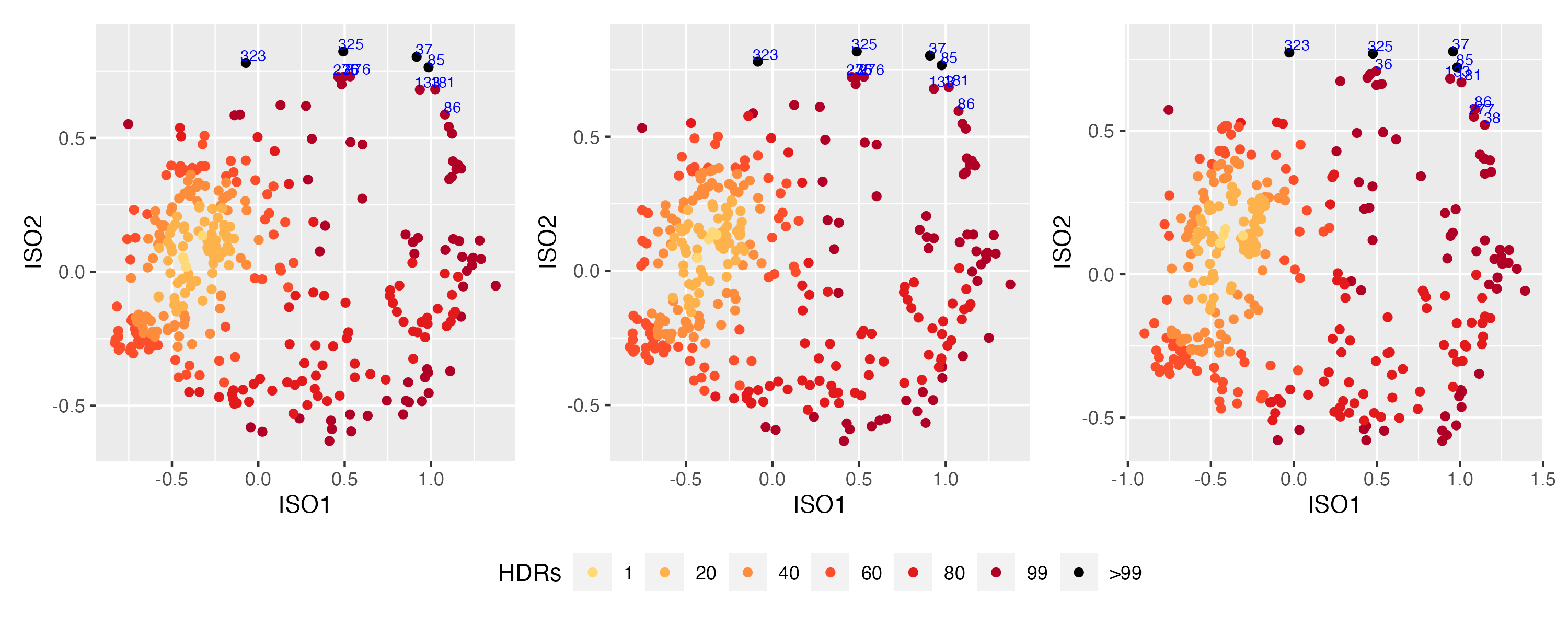} 

}

\caption{Highest density region plot for one household (Meter 1003) from the ISOMAP embeddings using exact NN (left panel), k-d trees (middle panel) and Annoy (right panel). The different colours represent highest density regions with different coverage probabilities. The most typical points (the 1\% HDR) are shown in yellow, while the top 10 most anomalous are shown in black with the time of week indexed in blue.}\label{fig:anomalies1id}
\end{figure}

One particular problem of interest in visualizing smart meter data is to
identify anomalous times of the week. For this reason,
\autoref{fig:anomalies1id} displays the same two-dimensional
representations from using ISOMAP and Annoy as \autoref{fig:todplot} but
with colors now used to demonstrate points in areas of high density
using the method of \citet{Hyndman1996-lk}. We identify the anomalies as
the observations with the top 10 lowest density values and these are
shown in black points with the time of week index labeled in blue. While
the anomalies identified by each manifold learning algorithm differ,
when the same manifold learning algorithm, ISOMAP, is used, the use of
approximate nearest neighbors has little impact on the identified
anomalies.

\begin{figure}

{\centering \includegraphics[width=1\linewidth]{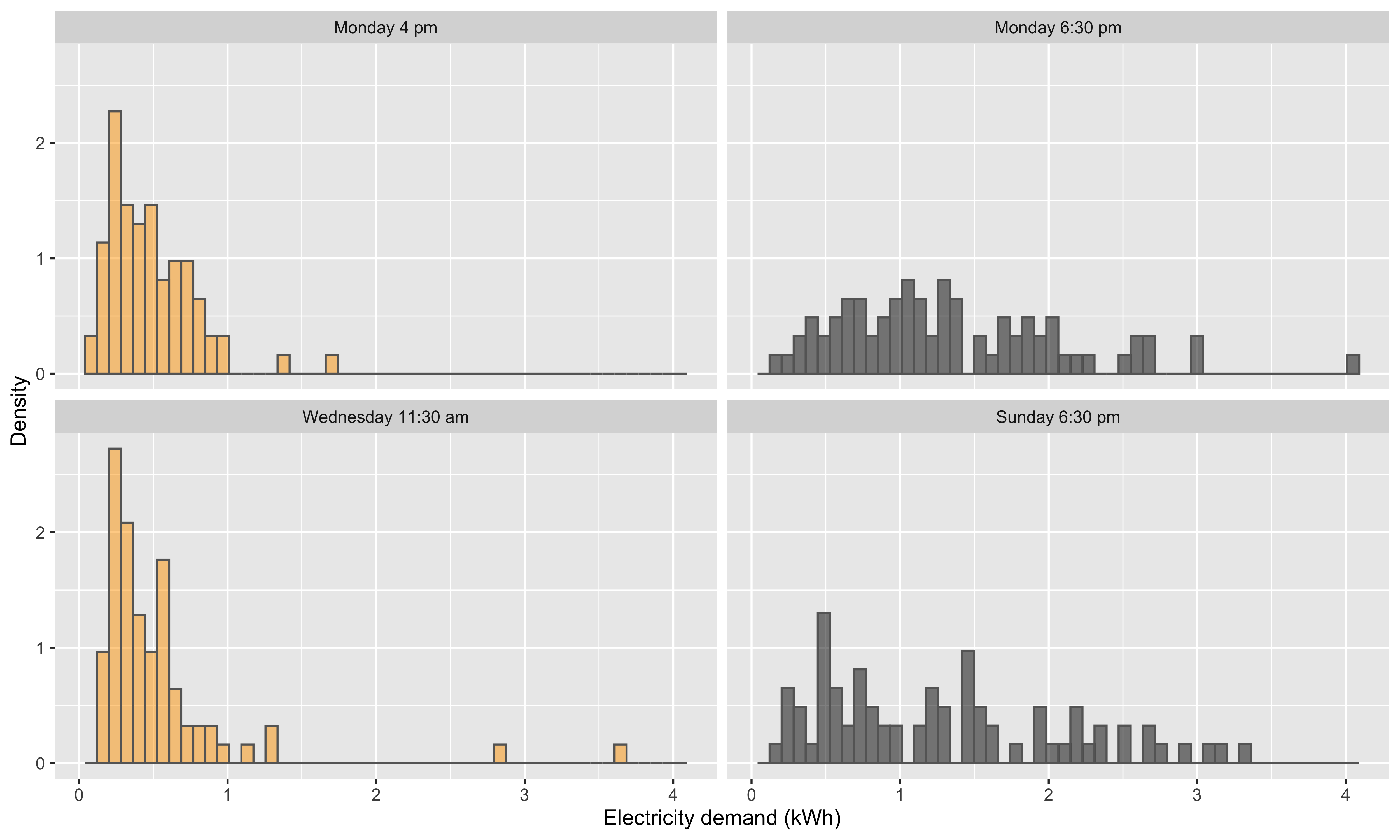} 

}

\caption{Electricity demand distribution plots for four time of week periods of meter ID 1003. The most typical periods, Monday 4pm and Wednesday 11:30am, are shown in orange (left panels), while the most anomalous periods, Monday 6:30pm and Sunday 6:30pm, are shown in black (right panels). }\label{fig:compare2tow}
\end{figure}

We can gain further insight by comparing distributions corresponding to
specific times of the week identified as anomalous and typical by
\autoref{fig:anomalies1id}. In \autoref{fig:compare2tow}, the left two
panels in orange correspond to the typical periods 32 and 119
(corresponding to Monday 4 pm and Wednesday 11:30 am respectively),
while the right two panels in black display the anomalous, periods 37
and 325 (corresponding to Monday 6:30 pm and Sunday 6:30 pm
respectively). The typical periods show relatively lower electricity
usage, and the distributions are skewed right. In contrast, for
anomalous periods, electricity demand is generally higher and
distributions exhibit a thicker right tail. In particular, there is a
more bimodal distribution on Sunday at 6:30 pm, which may suggest
certain patterns of behavior are undertaken by this specific household
on some but not all weeks at this time (e.g.~entertaining guests).

\hypertarget{comparison-between-households}{%
\subsection{Comparison between
households}\label{comparison-between-households}}

To compare all 3,639 households in the Irish smart meter data, we
consider two ways to compute distances between observations. The first
is to compute the density of each household's energy usage while
ignoring the time of the week with \(N=3,639\); using the notation of
\autoref{dataprocessing}, \(i\) corresponds to the household and \(t\)
corresponds to the half-hour period. However, since the time of week is
highly informative in electricity consumption data, an alternative would
be to compute densities such that the index \(i\) corresponds to a
household/time-of-week pair and index \(t\) corresponds to the week,
i.e.~\(N=3,639 \times 336=1,222,704\). In this case, we use \(F_{j,h}\)
to denote the distribution corresponding to household \(j\) and time of
week \(h\), and \(\pi_{j,k}\) to denote a vector whose entries give
values of a probability mass function of a discrete approximation to
\(F_{j,h}\). The distance between household \(j\) and household \(k\)
can then be computed as \begin{equation}\label{eq:betweenhhmetric}
  \delta_{j,k}=\sum\limits_{h=1}^{336}\hat{TV}(F_{j,h}\|F_{k,h}),
\end{equation} where \(\hat{TV}(\cdot,\cdot)\) is the total variation
metric between two discrete distributions. This metric resembles that
used by \citet{Hyndman2018-ia} who use a sum of Jenson-Shannon
divergences rather than the total variation metric. However, we propose
the use of total variation metric with an important advantage because
the \(\delta_{j,k}\) is equivalent to a Manhattan distance between the
stacked vectors \(\pi_j=(\pi'_{j,1},\ldots,\pi'_{j,336})'\) and
\(\pi_k=(\pi'_{k,1},\ldots,\pi'_{k,336})'\). Unlike
\citet{Hyndman2018-ia}, we do not have to compute all pairwise distances
(an \(O(N^2)\) operation) before computing nearest neighbors; instead,
we can compute exact nearest neighbors using k-d trees or, in an even
faster time, approximate nearest neighbors.

\begin{figure}

{\centering \includegraphics[width=1\linewidth]{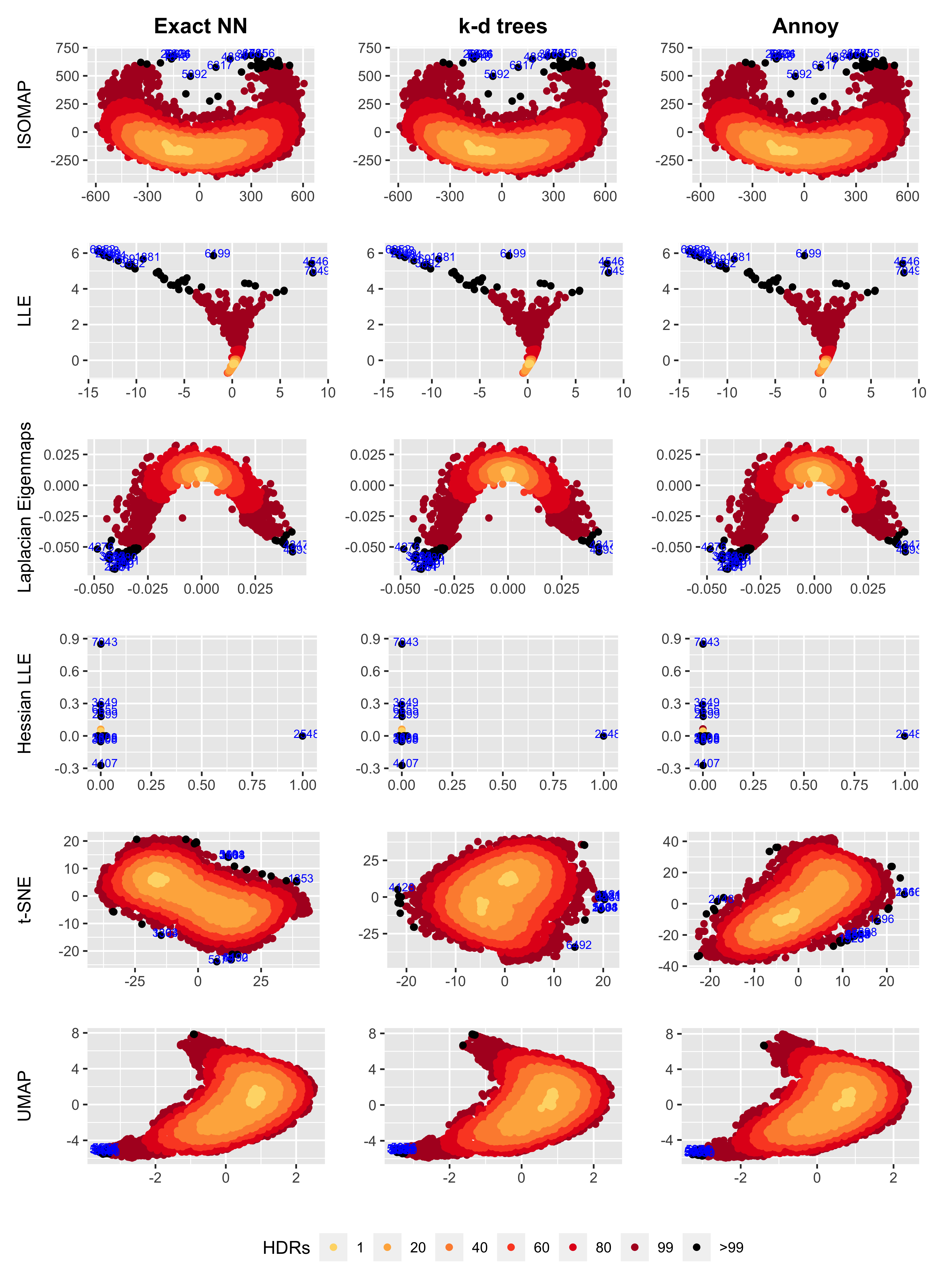} 

}

\caption{Highest density region plots from six manifold learning methods for all 3,639 households, with each point representing the distribution of one household. Top row: exact nearest neighbors. Middle row: ANN with k-d trees. Bottom row: ANN with Annoy.}\label{fig:allidhdr}
\end{figure}

\begin{table}

\caption{\label{tab:allidmeasure}Comparison of Trustworthiness measure for all household embeddings using true nearest neighbors, k-d trees, and Annoy for six manifold learning methods. ISOMAP and UMAP give the highest Trustworthiness while Hessian LLE gives the lowest. }
\centering
\resizebox{\linewidth}{!}{
\begin{tabu} to \linewidth {>{\raggedright}X>{}r>{}r>{}r>{}r>{}r>{}r}
\toprule
  & Isomap & LLE & Laplacian Eigenmaps & Hessian LLE & t-SNE & UMAP\\
\midrule
Exact NN & 0.926 & 0.911 & 0.880 & \textbf{0.579} & 0.914 & 0.943\\
ANN k-d trees & 0.926 & 0.911 & 0.880 & \textbf{0.579} & \textbf{0.919} & 0.943\\
ANN Annoy & \textbf{0.926} & \textbf{0.911} & \textbf{0.880} & 0.578 & 0.907 & \textbf{0.943}\\
\bottomrule
\end{tabu}}
\end{table}

\begin{table}

\caption{\label{tab:allidtime}Comparison of computation time for all household embeddings using true nearest neighbors, k-d trees, and Annoy for six manifold learning methods. Laplacian Eigenmaps with k-d trees saves the most computation time while Annoy is the slowest. }
\centering
\resizebox{\linewidth}{!}{
\begin{tabu} to \linewidth {>{\raggedright\arraybackslash}p{5cm}>{}r>{}r>{\raggedleft\arraybackslash}p{2cm}>{}r>{}r>{}r}
\toprule
  & Isomap & LLE & Laplacian Eigenmaps & Hessian LLE & t-SNE & UMAP\\
\midrule
Exact NN with brute-force & \textbf{4773.364} & \textbf{4967.419} & \textbf{4725.180} & \textbf{5930.408} & \textbf{4739.584} & 748.566\\
Exact NN with k-d trees & 755.810 & 1038.162 & 749.621 & 1204.143 & 747.696 & 744.262\\
ANN k-d trees & 755.684 & 1041.009 & 751.532 & 1208.720 & 747.825 & 741.715\\
ANN Annoy & 1670.209 & 1947.267 & 1656.094 & 2363.735 & 1557.124 & \textbf{1675.341}\\
\bottomrule
\end{tabu}}
\end{table}

Similar to the single household case in \autoref{electricityresults},
exact nearest neighbors are implemented by setting \(\varepsilon = 0\)
in k-d trees and approximate nearest neighbor is implemented using k-d
trees with \(\varepsilon = 1\) and Annoy with
\(\textit{n\_trees} = \textit{search\_k} = 50\). The distance metric
between households is defined as in Equation @ref(eq:betweenhhmetric).
For each scenario, the same six manifold learning algorithms are used as
in \autoref{electricityresults}. The highest density region plots of the
embeddings are shown in \autoref{fig:allidhdr} with exact nearest
neighbors on the left panel, ANN with k-d trees on the middle panel, and
ANN with Annoy on the right panel. Also, the trustworthiness measure is
computed for all combinations of manifold learning and nearest neighbor
searching algorithms with results summarized in
\autoref{tab:allidmeasure}, and the corresponding computation time is
reported in \autoref{tab:allidtime}.

ISOMAP, LLE, Laplacian Eigenmaps, and UMAP are highly robust to the use
of approximate nearest neighbors with the trustworthiness measures
equivalent across all nearest neighbor algorithms to the third decimal
place. Once again, ISOMAP, together with UMAP, yields fairly accurate
embedding according to the trustworthiness metric. Results are similar
if other accuracy metrics are used. The households identified as
anomalous do not change for ISOMAP, LLE, Laplacian Eigenmaps, and
Hessian LLE when approximate nearest neighbors are used. For the Hessian
LLE and t-SNE, the trustworthiness do change across different ANN
methods, however, the differences are minor. The Hessian LLE embedding
has the overall lowest trustworthiness measure which may explain why it
appears to exhibit some degeneracy in \autoref{fig:allidhdr}. An
alternative explanation for the poor performance of Hessian LLE is that
Hessian LLE is predicated on the existence of an isometric
\(d\)-dimensional embedding, and for this specific dataset such an
isometry may not exist for \(d=2\). As was the case in
\autoref{electricityresults}, whether the focus is upon the
identification of anomalies or the trustworthiness measure, the impact
of using different approximate nearest neighbor algorithms is
insubstantial when compared to the choice of the manifold learning
algorithm. However, in contrast to the earlier results,
\autoref{tab:allidtime} shows that a computational speedup is mostly
observed when an approximate version of k-d trees is used and not for
Annoy. This may be due to the fact that the discrete approximation is to
a multivariate density and hence has a higher dimensionality. Annoy is
generally recommended for data with a moderately high dimension of a few
hundred to a few thousand.

\begin{figure}

{\centering \includegraphics[width=0.8\linewidth]{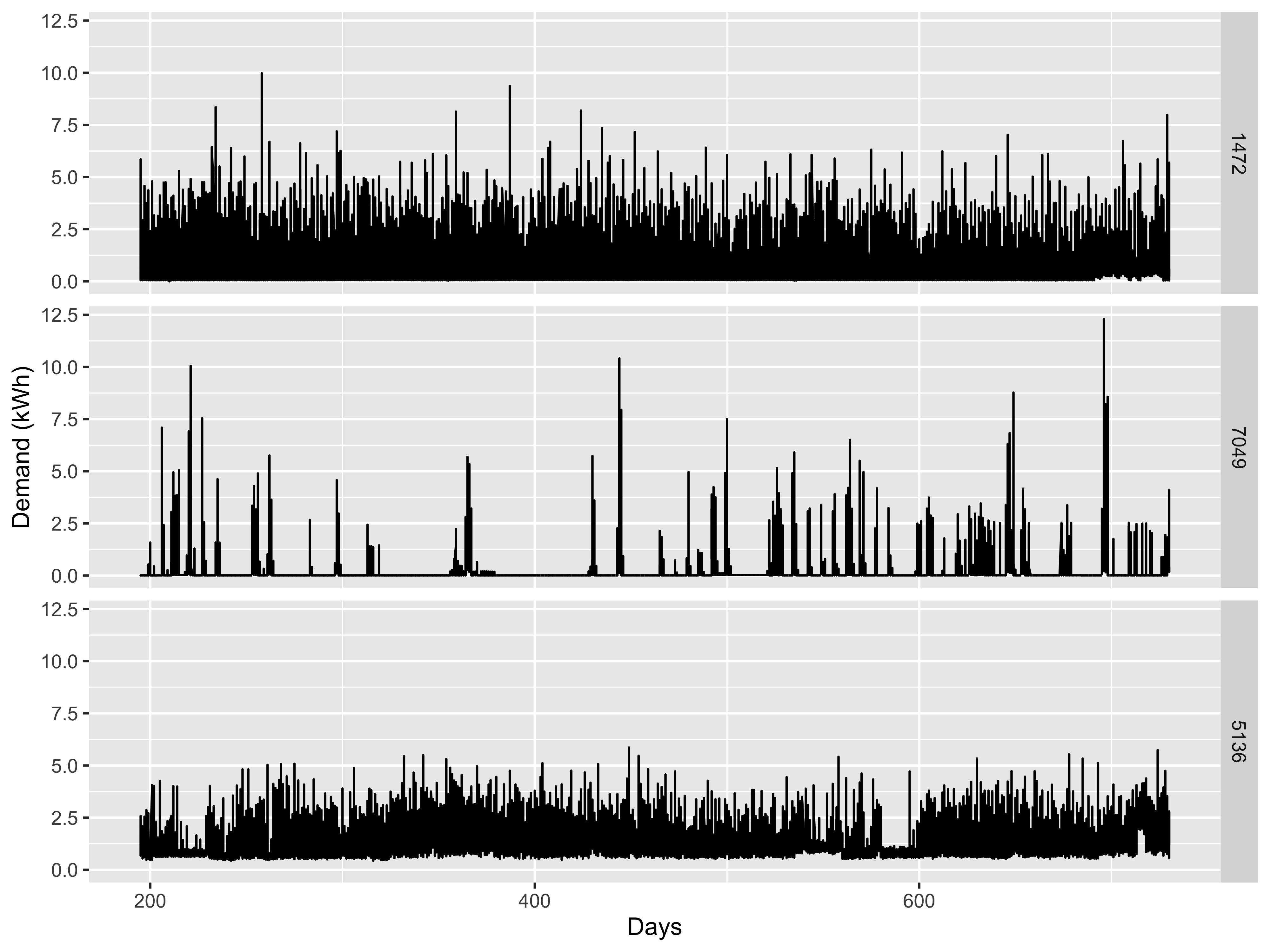} 

}

\caption{Electricity demand plots of all 535 days for one typical household 1472 and two anomalies, 7049 and 5136. }\label{fig:compare3ids}
\end{figure}
\begin{figure}

{\centering \includegraphics[width=1\linewidth]{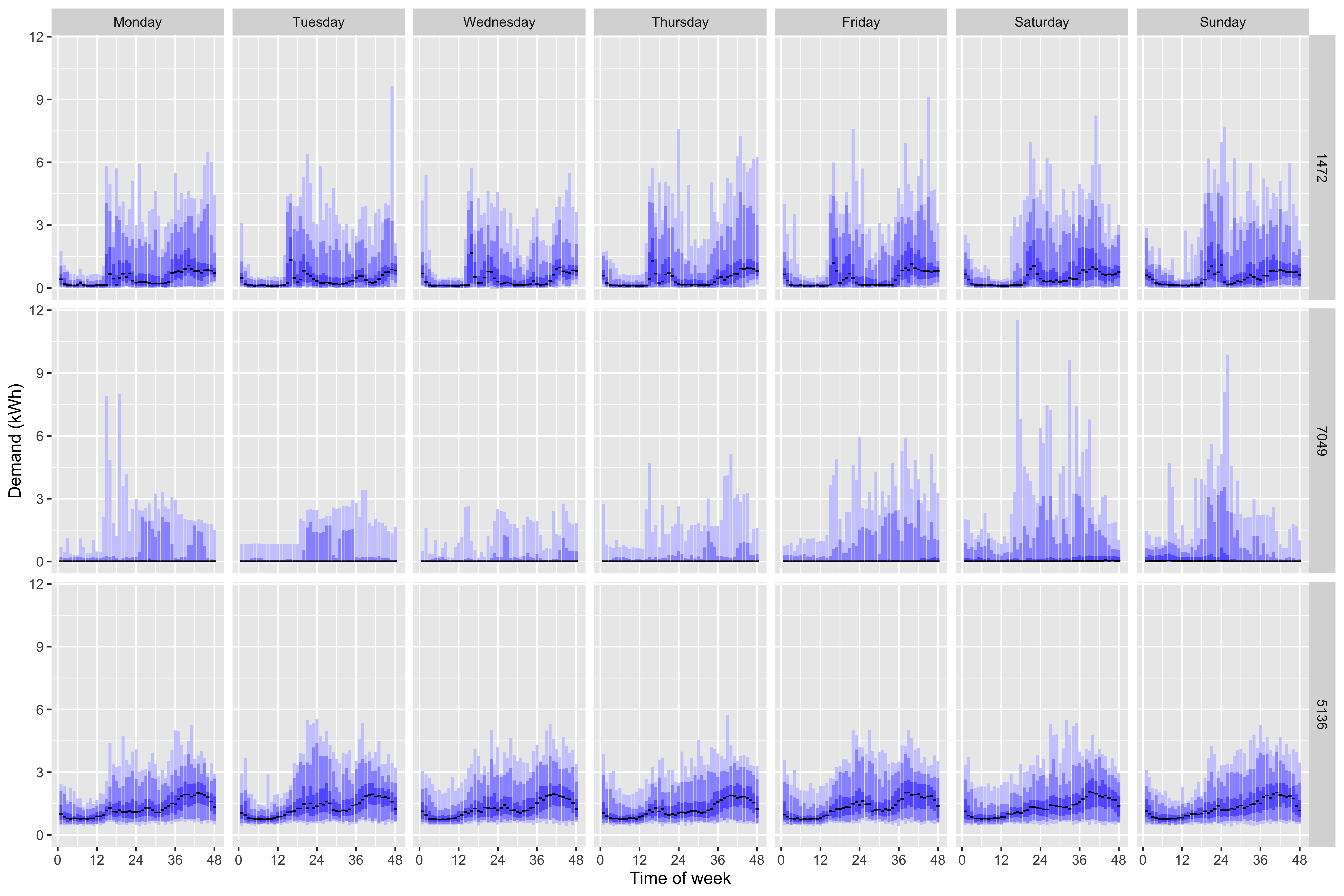} 

}

\caption{Quantile region plots of electricity demand against the time of week for one typical household 1472 and two anomalies, 7049 and 5136. The quantile regions displayed in the plot are 99.0\%, 95.0\%, 75\%, and 50\%. }\label{fig:hdrboxplot3ids}
\end{figure}

Further insights are gained by comparing the electricity demand
distributions of typical and anomalous households. For example, in the
ISOMAP highest density region plot, the meter IDs 7049 and 5136, are
both detected as anomalies. However, they lie far away from one other in
the embedding plot, suggesting that while both are anomalous, their
distributions are quite different. We can compare these to one typical
household, namely the household ID 1472). In \autoref{fig:compare3ids},
the electricity usage data of these three households are plotted over
all trial days, while the corresponding quantile region plots of
electricity demand by half-hour and day of the week are given in
\autoref{fig:hdrboxplot3ids}. For all three households, the time of the
week patterns are revealed in the top panel of
\autoref{fig:hdrboxplot3ids}. When compared to the typical household
1472, the electricity demand of the anomalous households is much lower.
The quantile region plots show that the distributions of the two
anomalous households, 7049 and 5136 are quite different from each other,
which is consistent with the relatively large distance between their
locations in the embedding plots. For household 7049, in most of the
days, the electricity demand is below 1 kWh per day indicating that
there are not many electrical applicants in this household or that
people are away for most of the trial days, which also explains some
spikes in certain days. The situation is quite different for household
5136 where for most of the time the electricity usage is below 3 kWh per
day and there are certain periods with almost no usage. This shows that
this household might have vacations for holidays and leave only the
necessary electrical applicants on while not at home. These findings
show that the manifold learning with approximate nearest neighbors works
well in finding anomalous households in the smart meter data despite the
large volume of data.

\hypertarget{conclusion}{%
\section{Conclusions}\label{conclusion}}

In this paper, we propose a new approximation method for statistical
manifold learning with consistent distance estimators. Based on i.i.d.
samples drawn from two probability distributions and a reference
distribution, Hellinger and total variation distance estimator are
constructed via a data-dependent partition technique. The estimators are
proved to be strongly consistent and could be linked to the \(L_2\) and
\(L_1\) norm where the vectors are built from empirical probabilities.
In doing so, rather than computing all pairwise Hellinger or total
variation distances, k-d trees and Annoy can be used to reduce
computational time.

We also explore the electricity usage patterns of different time periods
and households in the Irish smart meter dataset. In this case, the
elements of the manifold in question are probability distributions of
different time periods of a single household or different households.
Once again, we show that ANN techniques can be used within manifold
learning to save computation time and memory without having a severe
impact on the quality of the low-dimensional representations. For this
particular dataset, ISOMAP or UMAP together with k-d trees gives the
best tradeoff between embedding quality and computational time, while
Annoy breaks down for one example that is particularly high-dimensional.
Using the highest density region plots, we show how the techniques
developed can successfully identify both typical and anomalous times of
week and households.

There are several open questions to be explored. The first involves the
selection of tuning parameters for the approximate nearest neighbor
algorithm. The optimal choice may depend on a number of considerations;
for instance, if manifold learning is used to detect anomalies in an
online fashion, then the tuning parameters may be constrained in a way
that approximate nearest neighbors are found within a certain time frame
\citep{Talagala2020-ck}. Alternatively, tuning parameters may be
selected so that a maximal level of embedding quality is achieved, where
embedding quality is measured using one of the metrics we discuss in the
online supplementary document.

Finally, there remain a number of open issues in manifold learning
including the choice of intrinsic dimension \(d\) \citep{Denti2021-jl}
and the choice of manifold learning algorithm itself. In the case of the
former, we present results in \autoref{smartmeter} based only on \(d=2\)
due to our emphasis on visualizations and the ease of using
scatterplots. However, in principle, larger values of \(d\) could be
used and if visualization is required, the embedding dimensions could be
summarized using principal components or using projection pursuit
methods with random tours \citep{Cook1995-jx, Laa2020-qx}. Regarding the
choice of the manifold learning algorithm, we have seen that this has a
much bigger impact on accuracy measures than the use of approximate
nearest neighbors. This does bring into focus the need for further work
on the selection of manifold learning algorithms. However, it is an
encouraging result for the use of ANN which across a wide range of
algorithms improves the computational speed of manifold learning.

\hypertarget{supplementary-materials}{%
\section*{Supplementary Materials}\label{supplementary-materials}}
\addcontentsline{toc}{section}{Supplementary Materials}

The GitHub repository, \url{https://github.com/ffancheng/paper-mlann},
contains all materials required to reproduce this article. The code and
data files are also available online in the same repository. The online
supplementary,
\url{https://github.com/ffancheng/paper-mlann/blob/public/paper/OnlineSupplementary.pdf},
applies different techniques for manifold learning with approximate
nearest neighbors to the benchmark MNIST dataset and evaluates the
results with respect to computational time and embedding accuracy.

\hypertarget{acknowledgment}{%
\section*{Acknowledgment}\label{acknowledgment}}
\addcontentsline{toc}{section}{Acknowledgment}

This research was supported in part by the Monash eResearch Centre and
eSolutions-Research Support Services through the use of the MonARCH HPC
Cluster. We would also like to acknowledge the support of the
\emph{Australian Research Council (ARC) Centre of Excellence for
Mathematical and Statistical Frontiers (ACEMS)} for this project.

\bibliographystyle{agsm}
\bibliography{references.bib}

\clearpage\appendix
\section{Appendix A}

\subsection{Pseudocode for manifold learning algorithms}
\label{sec:mlalg}
This section provides the pseudocode for the four manifold learning algorithms in Section \ref{ml}.

\begin{algorithm}[!htb]
  \caption{ISOMAP}
  \label{alg:isomap}
  \DontPrintSemicolon
  \SetAlgoLined
  \SetKwInOut{Input}{Input}\SetKwInOut{Output}{Output}
  \Input{ high-dimensional data $x_i$ for all $i=1,\ldots,N$ }
  \Output{ low-dimensional embedding $y_i$ for all $i=1,\ldots,N$ }
  \BlankLine

  Construct the $K$-nearest neighbor graph $\mathcal{G}$ for $x_i$ where $i=1,\ldots,N$;

  Set edge weights to $\delta_{ij}$ for $x_i$ and $x_j$ connected by an edge in $\mathcal{G}$;

  Estimate the geodesic distances (distances along a manifold) between all pairs of points as the shortest-path distances on the graph $\mathcal{G}$;

  Using the estimates of the geodesic distances as inputs into classical MDS, obtain the output points $y_i$ for $i=1,\ldots,N$.

\end{algorithm}

\begin{algorithm}[!htb]
  \caption{LLE}
  \label{alg:lle}
  \DontPrintSemicolon
  \SetAlgoLined
  \SetKwInOut{Input}{Input}\SetKwInOut{Output}{Output}
  \Input{ high-dimensional data $x_i$ for all $i=1,\ldots,N$ }
  \Output{ low-dimensional embedding $y_i$ for all $i=1,\ldots,N$ }
  \BlankLine

  Construct the K-nearest neighbor graph $\mathcal{G}$ to find the $K$-ary neighborhood $U_K(i)$ of $x_i$ for all $i=1,\ldots,N$;

  Find a representation of each input point $x_i$ as a weighted and convex combination of its nearest neighbors by minimizing
  $$
    \bigg\|x_{i}-\sum_{j \in U_K(i)} w_{i j} x_{j}\bigg\|^{2}
  $$
  with respect to weights $w_{ij}$;

  Find a configuration that minimizes
  $$
    \sum_{i}\bigg\|y_{i}-\sum_{j \in U_K(i)} w_{i j} y_{j}\bigg\|^{2}
  $$
  with respect to $y_{1}, \dots, y_{n}$ where $w_{ij}$ are identical to those found in the previous step.

\end{algorithm}

\begin{algorithm}[!htb]
  \caption{Laplacian Eigenmaps}
  \label{alg:le}
  \DontPrintSemicolon
  \SetAlgoLined
  \SetKwInOut{Input}{Input}\SetKwInOut{Output}{Output}
  \Input{ high-dimensional data $x_i$ for all $i=1,\ldots,N$ }
  \Output{ low-dimensional embedding $y_i$ for all $i=1,\ldots,N$ }
  \BlankLine

  Construct the $K$-nearest neighborhood graph $\mathcal{G}$ for input points $x_i$ for all $i=1,\ldots,N$;

  For input points $x_i$ and $x_j$ that are connected by an edge on $\mathcal{G}$, compute weights as $w_{ij}=1$. Alternatively, set weights using the heat kernel $w_{i j}=e^{-\|x_{i}-x_{j}\|^{2} / 2 \sigma^{2}}$. For input points $x_i$ and $x_j$ that are not connected by an edge on $\mathcal{G}$, set $w_{ij}=0$;

  Compute the graph Laplacian as $L := D-W$, where $W$ has elements $w_{ij}$ and $D$ is the diagonal matrix with elements $D_{i i}=\sum_{j} w_{i j}$;

  Solve the generalized eigendecomposition equation $Lv = \lambda Dv$, where $\lambda$ and $v$ are the eigenvalue and eigenvector, respectively. The output points are obtained as the eigenvectors corresponding to the smallest non-zero eigenvalues.

\end{algorithm}

\begin{algorithm}[!htb]
  \caption{Hessian LLE}
  \label{alg:hlle}
  \DontPrintSemicolon
  \SetAlgoLined
  \SetKwInOut{Input}{Input}\SetKwInOut{Output}{Output}
  \Input{ high-dimensional data $x_i$ for all $i=1,\ldots,N$ }
  \Output{ low-dimensional embedding $y_i$ for all $i=1,\ldots,N$ }
  \BlankLine
  \begin{algorithmic}[1]

  \STATE Construct the $K$-nearest neighborhood graph $\mathcal{G}$;

  \FOR{$\ell = 1,\dots,N$}

  \STATE Estimate the tangent space at $x_\ell$ by stacking all $x_j:j\in U_K(\ell)$ as rows in a $K\times p$ matrix and take a singular value decomposition. A basis for the tangent space is determined by the first $d$ left singular vectors (i.e. a $K\times d$ matrix);

  \STATE Form a $K\times d(d+1)/2$ matrix, $Z$, by augmenting the $d$ left singular vectors from the previous step with a column of ones and with element-wise squares and cross-products of the $d$ left singular vectors;

  \STATE Conduct a Gram-Schmidt orthogonalization on $Z$ and transpose the result;

  \STATE Form the $d(d+1)/2\times K$ matrix $H^\ell$ by picking out the rows of the matrix from the previous step that correspond to the squared terms and cross products. Pre-multiplying a vector composed of elements $f(x_j):j\in U_K(\ell)$ by $H^\ell$ yields an estimate of the elements in the Hessian at $x_\ell$;

  \ENDFOR

  \STATE Compute an $N\times N$ discrete approximation of $\mathcal{H}(f)$ with entries
  $$
    \mathcal{H}_{i,j} = \sum_\ell \sum_r (H^\ell_{r,c(i)} H^\ell_{r,c(j)}),
  $$
  where $c(i)$ and $c(j)$ are the columns of $H^\ell$ corresponding to observation $i$ and $j$, and $r$ corresponds to an element of the Hessian;

  \STATE  Perform an eigendecomposition on the matrix from the previous step to approximate the null space of $\mathcal{H}$. This will have $d+1$ zero (or near zero) eigenvalues, the smallest of which corresponds to a constant $f$. The eigenvectors corresponding to the next $d$ smallest eigenvalues yield the output points.

  \end{algorithmic}
\end{algorithm}

\begin{algorithm}[!htb]
  \caption{t-SNE}
  \label{alg:tsne}
  \DontPrintSemicolon
  \SetAlgoLined
  \SetKwInOut{Input}{Input}\SetKwInOut{Output}{Output}\SetKwInOut{Parameter}{parameter}\SetKwInOut{OptParameter}{optimization parameter}
  \Input{ high-dimensional data $x_i$ for all $i=1,\ldots,N$ }
  \Output{ low-dimensional embedding $y_i$ for all $i=1,\ldots,N$ }
  \Parameter{ \textit{perplexity} }
  \OptParameter{ number of iterations $T$, learning rate $\eta$, momentum $\alpha(t)$ }
  \BlankLine
  \begin{algorithmic}[1]

  \STATE Compute the Gaussian conditional probability of distances for $x_i$ and $x_j$ as $p_{j \mid i}$ using
  $$
    p_{j \mid i}=\frac{\exp \left(-\left\|x_{i}-x_{j}\right\|^{2} / 2 \sigma_{i}^{2}\right)}{\sum_{k \neq i} \exp \left(-\left\|x_{i}-x_{k}\right\|^{2} / 2 \sigma_{i}^{2}\right)}
  $$
  where the variance $\sigma_{i}$ for each $x_i$ optimized with the perplexity parameter by $ 2 ^ {-\sum_j{p_{j \mid i} \log_2{p_{j \mid i}}}} = \textit{perplexity} $, and set $p_{i \mid i}=0$;

  \STATE Set the probability as $p_{i j}=(p_{j \mid i}+p_{i \mid j}) / (2 N)$;

  \STATE Initialize the optimization solution as $\mathcal{Y}^{(0)} = \{ y_1, \dots, y_N \}$ and $\mathcal{Y}^{(1)}$ sampled from normal distribution $\mathcal{N}(0, 10^{-4} \boldsymbol{I})$;

  \FOR{$t = 1,\dots,T$}

  \STATE Compute the Student t-distribution of distances between $y_i$ and $y_j$ as 
  $$
    q_{i j}=\frac{\left(1+\left\|y_{i}-y_{j}\right\|^{2}\right)^{-1}}{\sum_{k \neq l}\left(1+\left\|y_{k}-y_{l}\right\|^{2}\right)^{-1}};
  $$

  \STATE Compute the gradient of the Kullback-Leibler divergence loss function for all $i=1,\ldots,N$ as
  $$
    \frac{\delta KL}{\delta y_{i}}=4 \sum_{j}\left(p_{i j}-q_{i j}\right)\left(y_{i}-y_{j}\right)\left(1+\left\|y_{i}-y_{j}\right\|^{2}\right)^{-1} ;
  $$

  \STATE Update $ \mathcal{Y}^{(t)}=\mathcal{Y}^{(t-1)} + \eta \frac{\delta KL}{\delta \mathcal{Y}} + \alpha(t)\left(\mathcal{Y}^{(t-1)}-\mathcal{Y}^{(t-2)}\right). $

  \ENDFOR

  \end{algorithmic}
\end{algorithm}

\begin{algorithm}[!htb]
  \caption{UMAP}
  \label{alg:umap}
  \DontPrintSemicolon
  \SetAlgoLined
  \SetKwInOut{Input}{Input}\SetKwInOut{Output}{Output}\SetKwInOut{Parameter}{parameter}
  \Input{ high-dimensional data $x_i$ for all $i=1,\ldots,N$ }
  \Output{ low-dimensional embedding $y_i$ for all $i=1,\ldots,N$ }
  \Parameter{ number of nearest neighbors $K$, minimum distance between embedded points \textit{min\_dist}, amount of optimization work \textit{n\_epochs}}
  \BlankLine
  \begin{algorithmic}[1]

  \STATE Construct the $K$-nearest neighborhood graph to find the $K$-ary neighborhood $U_K(i)$ with distances $\delta_{ij}$, and denote the distance to its nearest neighbor as $\rho_i$ for each $x_i$;

  \STATE For the neighborhood points in $U_K(i)$, apply a smooth approximator to $\delta_{ij}$ to search for $\sigma_i$ such that $\sum_{i}^K e^{\frac{- \max(0, \delta_{ij} -\rho_{i}) } {\sigma_{i}}} = \log_2{K}$;

  \STATE Compute the exponential probability of distances between $x_i$ and $x_j$ as $p_{j \mid i}$ using
  $$
    p_{i \mid j}=e^{\frac{-\max(0, \delta_{ij} -\rho_{i})}{\sigma_{i}}};
  $$

  \STATE Set $p_{i j}=p_{j \mid i}+p_{i \mid j} - p_{i \mid j} p_{j \mid i}$;

  \STATE Define the distribution of distances between $y_i$ and $y_j$ as
  $$
    q_{i j}=\left(1+a\left(y_{i}-y_{j}\right)^{2 b}\right)^{-1},
  $$
  where $a$ and $b$ are found using the \textit{min\_dist} parameter such that
  $$
      q_{i j} \approx\left\{\begin{array}{ll}
  1 & \text { if } y_{i}-y_{j} \leq \textit{min\_dist} \\
  e^{-\left(y_{i}-y_{j}\right)-\textit{min\_dist}} & \text { if } y_{i}-y_{j} > \textit{min\_dist};
  \end{array}\right.
  $$

  \STATE The cost function is constructed using a binary fuzzy set cross entropy (CE) as
  $$
    CE(X, Y)=\sum_{i} \sum_{j}\left[p_{i j}(X) \log \left(\frac{p_{i j}(X)}{q_{i j}(Y)}\right)+\left(1-p_{i j}(X)\right) \log \left(\frac{1-p_{i j}(X)}{1-q_{i j}(Y)}\right)\right]
  $$

  \STATE Initialize the embedding coordinates $y_i$ using the graph Laplacian with weights as $p_{ij}$, similar to Laplacian Eigenmaps in Algorithm \ref{alg:le};

  \STATE Optimize $y_i$ by minimizing the cross entropy using stochastic gradient descent with \textit{n\_epochs}.

  \end{algorithmic}
\end{algorithm}

\clearpage
\FloatBarrier

\subsection{Pseudocode for approximate nearest neighbor searching methods}
\label{sec:annalg}
The pseudocode for three approximate nearest neighbor searching methods, k-d trees and Annoy are included in the following section.

The k-d trees method involves two steps: k-d tree construction in Algorithm \ref{alg:buildkdtree} and recursive searching in Algorithm \ref{alg:searchkdtree}.
Annoy also consists of two steps: preprocess in Algorithm \ref{alg:annoypre} and query in Algorithm \ref{alg:annoyquery}.

\begin{algorithm}[!htb]
  \caption{Constructing a k-d tree}
  \label{alg:buildkdtree}
  \begin{algorithmic}[1]
    \STATE Initialize the tree $\textit{depth}=0$ and the node index $g=0$;
    \STATE Initialize the root node $\mathcal{P}_g$ as $\mathbb{R}^p$;
    \STATE Initialize the set of all nodes at a depth of zero as $\mathcal{C}_0=\{\mathcal{P}_g\}$;
    \WHILE{$|\{x_i:x_i\in\mathcal{P}_h\}|>1$ for some $\mathcal{P}_h\in \mathcal{C}_{\textit{depth}}$}
      \STATE Set $\textit{depth}\leftarrow \textit{depth}+1$;
      \STATE Initialize $\mathcal{C}_{\textit{depth}}=\varnothing$;
      \FOR{$\mathcal{P}_h\in\mathcal{C}_{\textit{depth}-1}$}
        \STATE Find splitting dimension $\ell_h^*=\underset{\ell}{\textrm{argmax}}\underset{i,j\in\mathcal{P}_h}{\max}|x_{i\ell}-x_{j\ell}|$;
        \STATE Find splitting value $c_h^*=\big(x^{(\nu+1)}_{{\ell_h^*}}-x^{(\nu)}_{{\ell_h^*}}\big)/2$ where $x^{(r)}_{\ell}$ denotes $r^{th}$ order statistic of $\{x_i:x_{i}\in\mathcal{P}_h\}$ along dimension $\ell$;
        \STATE Set the partition for the left child node $\mathcal{P}^{\textit{left}}_h$ as the set of points $\{z:z\in\mathcal{P}_h, z_{\ell^*}<c^*\}$ where $z_{\ell^*}$ is the value of $z$ on the splitting dimension;
        \STATE Set the partition for the right child node $\mathcal{P}^{\textit{right}}_h$ as the set of points $\{z:z\in\mathcal{P}_h, z_{\ell^*}\geq c^*\}$;
        \STATE Set $\mathcal{P}_{g+1}:=\mathcal{P}_h^{\textit{left}}$ and $\mathcal{P}_{g+2}:=\mathcal{P}_h^{\textit{right}}$;
        \STATE Update $\mathcal{C}_{\textit{depth}}\leftarrow\mathcal{C}_{\textit{depth}}\cup\{\mathcal{P}_{g+1},\mathcal{P}_{g+2}\}$;
        \STATE Update $g\leftarrow g+2$;
      \ENDFOR
    \ENDWHILE
  \end{algorithmic}
\end{algorithm}

\begin{algorithm}[!htb]
  \caption{Recursive procedure $\mathtt{search}(\mathcal{P}_g)$ for nearest neighbor searching.}
  \label{alg:searchkdtree}
  \begin{algorithmic}[1]
  \IF{$\mathcal{P}_g$ is a terminal node}
  \STATE Compute distance $\delta_{iq}$ from query $x_q$ to $x_i\in\mathcal{P}_g$;
  \IF {$\delta_{iq}<\delta^*$}
  \STATE Set $\delta^*\leftarrow\delta$ and $x_{q^*}\leftarrow x_{i}$;
  \ENDIF
  \RETURN
  \ELSE
  \STATE For splitting dimension $l_g^*$ and splitting value $c_g^*$,
  \IF{$x_{q{l_g^*}}<c_g^*$}
  \STATE Evaluate $\mathtt{search}(\mathcal{P}^{\textit{left}}_g)$;
  \STATE Find hyper-sphere $\mathcal{S}$ of radius $\delta^*$ and tightest bounding box $\mathcal{B}$ of $\{x_i:x_i\in\mathcal{P}^{\textit{right}}_g\}$;
  \IF{$\mathcal{S}$ and $\mathcal{B}$ intersect}
  \STATE Evaluate $\mathtt{search}(\mathcal{P}^{\textit{right}}_g)$;
  \ENDIF
  \RETURN
  \ELSE
  \STATE Evaluate $\mathtt{search}(\mathcal{P}^{\textit{right}}_g)$;
  \STATE Find hyper-sphere $\mathcal{S}$ of radius $\delta^*$ and tightest bounding box $\mathcal{B}$ of $\{x_i:x_i\in\mathcal{P}^{\textit{left}}_g\}$;
  \IF{$\mathcal{S}$ and $\mathcal{B}$ intersect}
  \STATE Evaluate $\mathtt{search}(\mathcal{P}^{\textit{left}}_g)$;
  \ENDIF
  \RETURN
  \ENDIF
  \ENDIF
  \end{algorithmic}
\end{algorithm}

\begin{algorithm}[!b]
  \caption{Annoy preprocess}
  \label{alg:annoypre}
  \begin{algorithmic}[1]
    \STATE Initialize the node index $g=0$ and tree $\textit{depth}=0$;
    \STATE Initialize each root node as $\mathcal{P}_{0,t}=\mathbb{R}^p$ for $t=1,\dots,\textit{n\_trees}$;
    \STATE Initialize $\mathcal{C}_{0,t}=\{\mathcal{P}_{0,t}\}$ for $t=1,\dots,\textit{n\_trees}$;
    \FOR{$t = 1,\dots, \textit{n\_trees}$}
      \WHILE{$|\{x_i:x_i\in\mathcal{P}_{h,t}\}|>\kappa$ for some $\mathcal{P}_{g,t}\in \mathcal{C}_{\textit{depth},t}$}
        \STATE Set $\textit{depth}\leftarrow \textit{depth}+1$
        \STATE Initialize $\mathcal{C}_{\textit{depth},t}=\varnothing$
        \FOR{$\mathcal{P}_{h,t}\in \mathcal{C}_{\textit{depth}-1,t}$}
        \STATE Randomly select two points $x_i,x_j\in\mathcal{P}_{h,t}$;
        \STATE Find the hyperplane $\mathcal{S}_{h,t}$ equidistant from $x_i$ and $x_j$;
        \STATE Set the left child node partition $\mathcal{P}^{\textit{left}}_{h,t}$ as all points $\{z:z\in\mathcal{P}_{h,t},\delta_{zi}<\delta_{zj}\}$ where $\delta_{zi}$ and $\delta_{zj}$ are the distances from $z$ to $x_i$ and $x_j$ respectively;
        \STATE Set the right child node partition $\mathcal{P}^{\textit{right}}_{h,t}$ as the set of all points $\{z:z\in\mathcal{C}_g,\delta_{zi}\geq \delta_{zj}\}$;
        \STATE Update $\mathcal{P}_{g+1,t}:=\mathcal{P}_{g,t}^{\textit{left}}$ and $\mathcal{P}_{g+2,t}:=\mathcal{P}_{g,t}^{\textit{right}}$;
        \STATE Update $\mathcal{C}_{\textit{depth},t}\leftarrow\mathcal{C}_{\textit{depth},t}\cup\{\mathcal{P}_{g+1,t},\mathcal{P}_{g+2,t}\}$
        \STATE Update $g\leftarrow g+2$;
        \ENDFOR
      \ENDWHILE
    \ENDFOR
  \end{algorithmic}
\end{algorithm}

\begin{algorithm}[!b]
  \caption{Annoy query}
  \label{alg:annoyquery}
  \begin{algorithmic}[1]
    \STATE Letting, $\mathcal{S}_{0,t}$ be the hyperplane splitting the root node of tree $t$, find $\mathcal{P}^q_{0,t}$ and $\mathcal{P}^{-q}_{0,t}$ where the former is on the same side of $\mathcal{S}_{0,t}$ as $x_q$ latter is on the opposite side of $\mathcal{S}_{0,t}$ from $x_q$;
    \STATE Set $\mathcal{C}_0\leftarrow\{\mathcal{P}^q_{0,t}\}$ for $t=1,\dots,\textit{n\_trees}$;
    \STATE Set up a priority queue $\mathcal{Q}$ made up of pairs $(\mathcal{P}_g,\delta_g)$ where $\mathcal{P}_g$ is a partition of space (initialized at $\mathcal{P}^{-q}_{0,t}$ for all $t$) and $\delta_g$ are the distances from $x_q$ to the hyperplane corresponding to each $\mathcal{P}_g$;
    \STATE Set $\textit{depth}\leftarrow 0$;
    \WHILE{alsot least one element of $\mathcal{C}_{\textit{depth}}\cup\mathcal{Q}_{\textit{depth}}$ does not correspond to a leaf node}
    \FOR{$\mathcal{P}_{g}\in\mathcal{Q}$}
    \STATE Find $\mathcal{P}^q_g$ and $\mathcal{P}^{-q}_g$ where these are defined in a similar fashion to Step 1;
    \STATE Replace $\mathcal{P}_g$with $\mathcal{P}^q_g$ in the priority queue;
    \STATE Add $\mathcal{P}^{-q}_g$ to the priority queue;
    \IF{$|\mathcal{Q}|>maxsize\_queue$}
    \STATE Remove $\mathcal{Q}_{maxsize\_queue}$ from $\mathcal{Q}$;
    \ENDIF
    \ENDFOR
    \STATE Set $\textit{depth}\leftarrow \textit{depth}+1$, $C_{\textit{depth}}\leftarrow\varnothing$;
    \FOR{$\mathcal{P}_h\in\mathcal{C}_{\textit{depth}-1}$}
    \STATE Find $\mathcal{P}^q_h$ and $\mathcal{P}^{-q}_h$ where these are defined in a similar fashion to Step 1;
    \STATE Set $\mathcal{C}_{\textit{depth}}\leftarrow\mathcal{C}_{\textit{depth}}\cup\mathcal{P}^q_h$;
    \STATE Add $\mathcal{P}_h^{-q}$ to $\mathcal{Q}$, removing elements in the same fashion as Steps 10-12 if needed;
    \ENDFOR
    \ENDWHILE
    \STATE Set candidate set $\mathcal{K}\leftarrow\underset{\mathcal{P}_h\in\mathcal{C}_{\textit{depth}}}{\cup}x_i\in\mathcal{P}_h$;
    \STATE Set $g\leftarrow 0$;
    \WHILE{$|\mathcal{K}|<search\_k$}
    \STATE Set $g\leftarrow g+1$;
    \STATE Set $\mathcal{K}\leftarrow\mathcal{K}\cup\{x_i:x_i\in\mathcal{P}_g\}$ where $\mathcal{P}_g$ is the $g^{th}$ element of the priority queue;
    \ENDWHILE
    \STATE Search for nearest neighbors among the elements of $\mathcal{K}$ by brute force;
  \end{algorithmic}
\end{algorithm}


\clearpage

\end{document}